\documentclass{article} 
\usepackage[final]{colm2025_conference}

\usepackage{microtype}
\usepackage{hyperref}
\usepackage{url}
\usepackage{booktabs}

\usepackage{lineno}

\definecolor{darkblue}{rgb}{0, 0, 0.5}
\hypersetup{colorlinks=true, citecolor=darkblue, linkcolor=darkblue, urlcolor=darkblue}

\usepackage{graphicx}
\usepackage{amsmath}
\usepackage{listings}
\usepackage[most]{tcolorbox}
\usepackage{booktabs}
\usepackage{multirow}
\usepackage{fvextra}
\usepackage{wrapfig}
\usepackage{graphicx}
\usepackage{subcaption}
\usepackage{array}
\usepackage{xspace}

\newcommand{\benchmark}{\textsc{Cupid}}
\newcommand{\emoji}{\includegraphics[height=0.95em,trim=0 .4em 0 0]{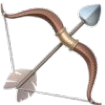}}
\newcommand{\benchmarkWithEmoji}{\emoji~\benchmark{}}
\newcommand{\matcherName}{\textsc{PrefMatcher-7B}}
\newcommand{\promptbox}[3]
{
    \begin{tcolorbox}[width=\linewidth, fontupper=\tiny]
        \textbf{#1} \\  

        \hrule
        \bigskip

        \textbf{System Prompt} \\

        \begin{Verbatim}[breaklines, fontsize=\tiny]
#2
        \end{Verbatim}

        \hrule
        \bigskip

        \IfNoValueF{#3}{ 
            \textbf{User Prompt} \\

            \begin{Verbatim}[breaklines, fontsize=\tiny]
#3
            \end{Verbatim}
        }
    \end{tcolorbox}
}
\newcommand{\github}{\raisebox{-1.5pt}{\includegraphics[height=1.05em]{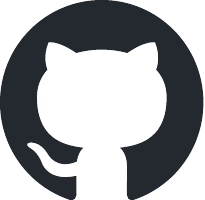}}\xspace}
\DeclareSymbolFont{extraup}{U}{zavm}{m}{n}
\DeclareMathSymbol{\varheart}{\mathalpha}{extraup}{86}
\DeclareMathSymbol{\vardiamond}{\mathalpha}{extraup}{87}

\newcommand\kaist{$^\heartsuit$}
\newcommand\snu{$^{\spadesuit}$}
\newcommand\calvin{$^{\diamondsuit}$}
\newcommand\naver{$^{\clubsuit}$}
\newcommand\aspace{~~}

\title{\benchmarkWithEmoji{}: Evaluating Personalized and Contextualized \\ Alignment of LLMs from Interactions}

\author{Tae Soo Kim\kaist\aspace
     Yoonjoo Lee\kaist\aspace
     Yoonah Park\snu\aspace
     Jiho Kim\calvin\aspace \\ 
     \textbf{Young-Ho Kim}\naver\aspace
     \textbf{Juho Kim}\kaist\aspace \\
     \kaist{}~KAIST \aspace 
     \snu{}~Seoul National University \aspace
     \calvin{}~Calvin University \aspace
     \naver{}~NAVER AI LAB \aspace \\
     \texttt{taesoo.kim@kaist.ac.kr}
}

\begin{document}

\ifcolmsubmission
\linenumbers
\fi

\vspace{-16pt}

\maketitle

\vspace{-20pt}

\begin{abstract}
\vspace{-8pt}
Personalization of Large Language Models (LLMs) often assumes users hold static preferences that reflect globally in all tasks.
In reality, humans hold dynamic preferences that change depending on the context.
As users interact with an LLM in various contexts, they naturally reveal their contextual preferences, which a model must infer and apply in future contexts to ensure alignment.
To assess this, we introduce \benchmarkWithEmoji{}, a benchmark of 756 human-curated interaction session histories between users and LLM-based chat assistants.
In each interaction session, the user provides a request in a specific context and expresses their preference through multi-turn feedback.
Given a new user request and prior interaction sessions, our benchmark assesses whether LLMs can infer the preference relevant to this request and generate a response that satisfies this preference.
With \benchmark{}, we evaluated 10 open and proprietary LLMs, revealing that state-of-the-art LLMs struggle to infer preferences from multi-turn interactions and fail to discern what previous context is relevant to a new request---under 50\% precision and 65\% recall.
Our work highlights the need to advance LLM capabilities for more contextually personalized interactions and proposes \benchmark{} as a resource to drive these improvements.

\begin{center}
    \renewcommand{\arraystretch}{1.2}
    \begin{tabular}{rl}
         \github & \href{https://github.com/kixlab/CUPID}{\path{https://github.com/kixlab/CUPID}} \\
    \end{tabular}
\end{center}

\end{abstract}
\begin{wrapfigure}{r}{0.449\textwidth}
    \vspace{-44pt}
    \begin{center}
    \includegraphics[width=0.449\textwidth]{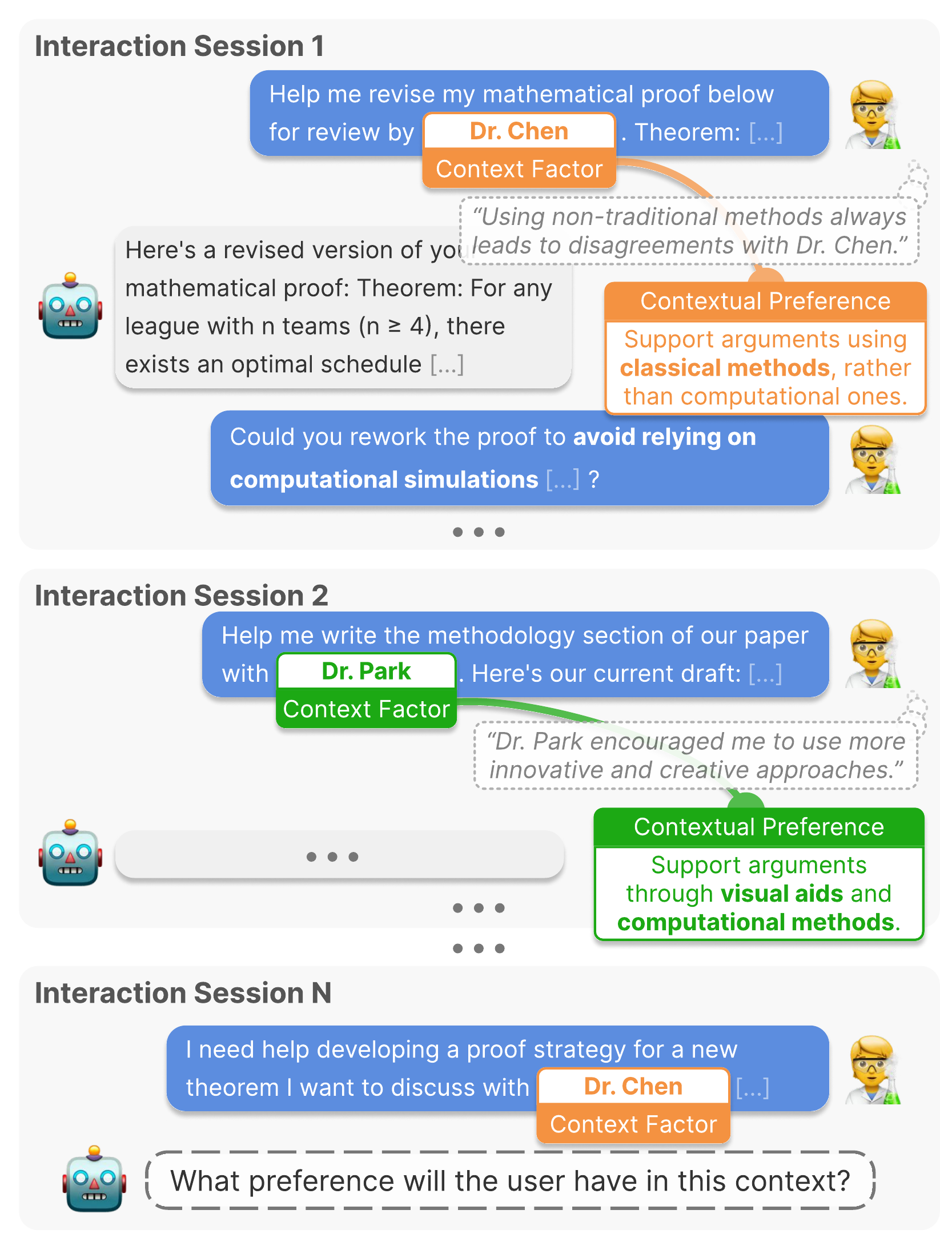}
    \vspace{-20pt}
    \caption{Example instance in \benchmarkWithEmoji{} illustrates a user that holds distinct preferences in different contexts due to personal experiences, where the preferences are only revealed to the LLM through the user's feedback in prior interactions.}
    \label{fig:teaser}
    \vspace{-20pt}
    \end{center} 
\end{wrapfigure}

\vspace{-12pt}
\section{Introduction}

\vspace{-4pt}

Large Language Models (LLMs) have shown remarkable capabilities across various tasks~\citep{nakano2021webgpt, achiam2023gpt}, benefiting users through diverse applications and conversational assistants~\citep{openai2022chatgpt, anthropic2023claude, microsoft2023copilot}.
Aligning these models with human values and preferences is crucial as they are increasingly integrated into user experiences~\citep{ji2023ai}.
Initial efforts focused on aligning LLMs on broad, general values (e.g., helpfulness, harmlessness, honest)~\citep{bai2022training, bai2022constitutional} or the aggregated preferences of diverse users~\citep{ouyang2022training, kopf2024openassistant}.
Recognizing that these approaches overlook users' diverse expectations, more recent work focuses on \textit{personalized alignment}, where LLMs are aligned with the users' individual preferences~\citep{wu2024aligning, jang2023personalized, lee2024aligning}.
Despite this progress, existing work largely assumes that users have \textit{static} preferences (i.e., preferences are global and do not vary over time)~\citep{wu2024aligning, lee2024aligning} or focus on stable characteristics (e.g., sociodemographics) to characterize  preferences~\citep{kirk2024prism}.

In reality, human preferences are \textit{context-dependent}~\citep{warren2011values, baker2009action, malaviya2024contextualized}: an individual's intents and expectations shift depending on their situation.
For example, in \autoref{fig:teaser}, a researcher consults an LLM to refine their paper's writing but their preference varies by collaborator: focus on classical methods with Dr. Chen due to prior disagreements while embracing computational methods with Dr. Park who advocated for them.
These contextual variability means that models must understand what values and preferences a user holds in distinct contexts.
Interactions between a user and an LLM may indirectly reveal these shifting preferences as users provide feedback to the model in diverse situations~\citep{wu2024aligning, kim2024understanding}.
However, it is unclear whether current LLMs possess the capability to identify a user's preference from their feedback, infer the relevant context, and proactively apply this knowledge in future interactions.

In this work, we present \benchmarkWithEmoji{}\footnote{~\url{https://huggingface.co/datasets/kixlab/CUPID}}
(\textbf{C}ontextual \textbf{U}ser \textbf{P}reference \textbf{I}nference \textbf{D}ataset),
a challenging and scalable benchmark that is designed to assess LLMs' ability to infer users' preferences tied to diverse contexts from user-LLM interaction histories.
\benchmark{} consists of 756 human-curated interaction histories, where each history presents a series of \textit{interaction sessions} or dialogues between a simulated user and an LLM.
Each interaction session presents a task-oriented dialogue where a user asks a request explicitly stating the context, and then gradually reveals their contextual preference through multi-turn feedback to the LLM.
Given a new request and a history of previous interaction sessions, the benchmark evaluates whether LLMs can (1) \textit{infer} the user's preference related to the context of the new request, and (2) \textit{generate} a response to the request that will satisfy this preference.

To construct \benchmark{}, we designed a pipeline that generates diverse and rich interaction sessions.
After generating a persona pool, for each persona, the pipeline generates a list of \textit{context factors} (i.e., significant people, objects, locations, etc. in the persona's world that influences their expectations) and \textit{contextual preference} (i.e., value, principle, or criterion associated with each factor).
For each persona, the pipeline then generates a series of interaction sessions that involve these factors and preferences.
We combine the concepts of LLM-as-a-Judge~\citep{zheng2023judging, ye2023flask} and LLM-simulated users~\citep{wu2024longmemeval, wu2024aligning} to create multi-turn dialogues where a simulated user evaluates and provides feedback to LLM's responses, gradually disclosing its contextual preferences.
All the data was verified with human annotators and $\sim${}9\% of the data was manually edited.

With \benchmark{}, we assess 10 open and proprietary LLMs.
We find that \textit{all models} struggle to infer users' preferences based on past interaction sessions, with no model exceeding 50\% precision and 65\% recall.
Models failed at recognizing relevant contexts in prior interactions and extracting preferences from multi-turn conversations.
We find a strong correlation (r=0.764) between performance in inferring preferences and generating responses that satisfy them, suggesting that the capability to infer context-dependent preferences underpins personalization capabilities.
Finally, we present a finetuned metric, \matcherName{}\footnote{~\url{https://huggingface.co/kixlab/prefmatcher-7b}}, to reduce the cost of evaluation on our benchmark and a larger unverified dataset\footnote{~\url{https://huggingface.co/datasets/kixlab/CUPID-Unverified}} to support training and further research.
\vspace{-6pt}
\section{\benchmarkWithEmoji{} Benchmark}

\vspace{-4pt}
\benchmark{} evaluates LLMs' ability to infer a user's distinct preference in different contexts from prior interactions between the user and the LLM.
\vspace{-6pt}

\subsection{Definitions}
\vspace{-4pt}

Our benchmark is composed of interaction sessions $S_i = (c_i, p_i, D_i)$.
\vspace{-4pt}

\paragraph{Context Factor, $c_i$} 
Represents an element (e.g., person, location, tool, activity) in a user's environment or world, which plays a role in and influences that interaction session.
We assume that each session's context is only defined and influenced by a single factor.
\vspace{-6pt}

\paragraph{Contextual Preference, $p_i$} 
Represents a value, principle, criterion, requirement, or constraint that the user holds when the context factor $c_i$ is involved in the situation.
While prior work focused on more simple and concrete lifestyle preferences (e.g., \textit{"I cannot eat spicy food"})~\citep{wu2024aligning, zhao2025llms}, we focus on more complex, abstract, and task-oriented preferences (e.g., \textit{"Instructions must break down complex procedures into numbered micro-steps with verification points after each major section"})---more examples in \autoref{tab:dataset-examples}.
Preferences in our dataset often require multiple rounds of feedback to be satisfied.
\vspace{-6pt}

\paragraph{Dialogue, $D_i = (u_{i,1},m_{i,1},...,m_{i,L_i-1},u_{i,L_i})$}
Represents an interaction between a user and an LLM. 
The dialogue begins with the user's initial request $u_{i,1}$ that explicitly states the context factor $c_i$. In subsequent turns, the user iteratively provides feedback until the model's responses $(m_1, m_2, ..., m_{i,L_i-1})$ satisfy preference $p_i$. The dialogue length $L_i$ is determined when $p_i$ is fully satisfied, confirmed by the user’s final utterance $u_{i,L_i}$.

\vspace{-2pt}
\subsection{Problem Formulation}

Each task instance in \benchmark{} consists of a 4-tuple $(u_\text{current}, c_\text{current}, p_\text{current}, \mathbf{S})$, where $u_\text{current}$ represents the user's current request to the LLM, $c_\text{current}$ and $p_\text{current}$ represent the context factor and contextual preference in the new request, and $\mathbf{S}$ is the history of previous interaction sessions, where at least one $S_i \in \mathbf{S}$ satisfies $c_i = c_\text{current}$ and $p_i = p_\text{current}$.
Our benchmark evaluates LLMs on two tasks:
\vspace{-4pt}
\begin{itemize}
    \item \textbf{Inference}: Given $u_\text{current}$ and $\mathbf{D} = \{D_i \mid S_i \in \mathbf{S}\}$, the model should infer the user's contextual preference $p_\text{current}$.
    \item \textbf{Generation}: Given $u_\text{current}$ and $\mathbf{D} = \{D_i \mid S_i \in \mathbf{S}\}$, the model should generate a response $r$ that will satisfy or align with the user's preference $p_\text{current}$.
\end{itemize}

\vspace{-4pt}
\subsection{Metrics}
\label{sec:metrics}

\vspace{-4pt}
\paragraph{Inference - Preference Match}
Preferences in our benchmark are complex, with each preference entailing multiple fine-grained expectations (i.e., sub-preferences). 
An inferred preference should capture all sub-preferences (\textit{recall}) without including irrelevant ones (\textit{precision}).
To assess this, we design an LLM-based \textit{preference matching} metric, inspired by work on measuring factual precision through atomic facts~\citep{min2023factscore} and fine-grained checklist evaluations~\citep{lin2024wildbench, cook2024ticking}.
Specifically, we first use an LLM to \textit{decompose} the ground-truth and inferred preferences, $p$ and $\hat{p}$, into \textit{atomic checklists} $Q_p = \{q_1, ..., q_n\}$ and $Q_{\hat{p}} = \{\hat{q}_1, ..., \hat{q}_n\}$, where each checklist item $q_i$ or $\hat{q}_i$ assesses a single sub-preference.
Then, we employ an LLM with a few-shot prompt (\autoref{fig:preference-match-prompt}) to evaluate whether each checklist item \textit{matches} the other preference:
\vspace{-1pt}
\[
\textsc{Match}(\hat{q}, p)
=
\begin{cases}
1, & \text{if aligning with $p$ fully covers $\hat{q}$,}\\
0.5, & \text{if aligning with $p$ partially covers $\hat{q}$,}\\
0, & \text{otherwise}.
\end{cases}
\]

\vspace{-4pt}
We use GPT-4o for decomposing and matching.
For matching, we conducted a meta-evaluation with 230 human-annotated data points, separate from our dataset.
GPT-4o achieved a Krippendorff's alpha~\citep{krippendorff2004reliability} of 0.769 (\textit{substantial agreement}) with the majority vote of human annotators.
To reduce the cost of evaluating on our benchmark, we also present \matcherName{}, a finetuned model for preference matching that achieves a Krippendorff's alpha of 0.748 (\textit{substantial agreement}).
Prompts, human meta-evaluation details, and finetuning details are in Appendix~\ref{appendix:metrics-match}.
Finally, we compute: $\mathrm{Precision} = \frac{1}{|Q_{\hat{p}}|}\sum_{\hat{q} \in Q_{\hat{p}}}\textsc{Match}(\hat{q}, p)$, 
$\mathrm{Recall} = \frac{1}{|Q_p|}\sum_{q \in Q_p}\textsc{Match}(q, \hat{p})$, and 
the F1 score.

\vspace{-2pt}
\paragraph{Generation - Preference Alignment}
For the generation task, we evaluate a model's response $r$ on its alignment with the contextual preference $p_\text{current}$ using the \textit{LLM-as-a-Judge} approach~\citep{zheng2023judging}. 
As prior work has shown that LLMs can evaluate other models' responses on diverse skills, principles, or criteria~\citep{fu2023gptscore, kim2024evallm, ye2023flask}, we prompt GPT-4o to provide a score ranging from 1 to 10 on the degree to which a model response satisfies the ground-truth contextual preference.
As checklists can increase the consistency and reliability of LLM evaluations~\citep{lin2024wildbench, cook2024ticking}, we also provide the automatically decomposed checklist $Q_p$ used in preference matching to the LLM judge.
Prompt and details are in Appendix~\ref{appendix:metrics-quality}.

\begin{figure*}[b!] \begin{center}
    \vspace{-8pt}
    \includegraphics[width=0.95\linewidth]{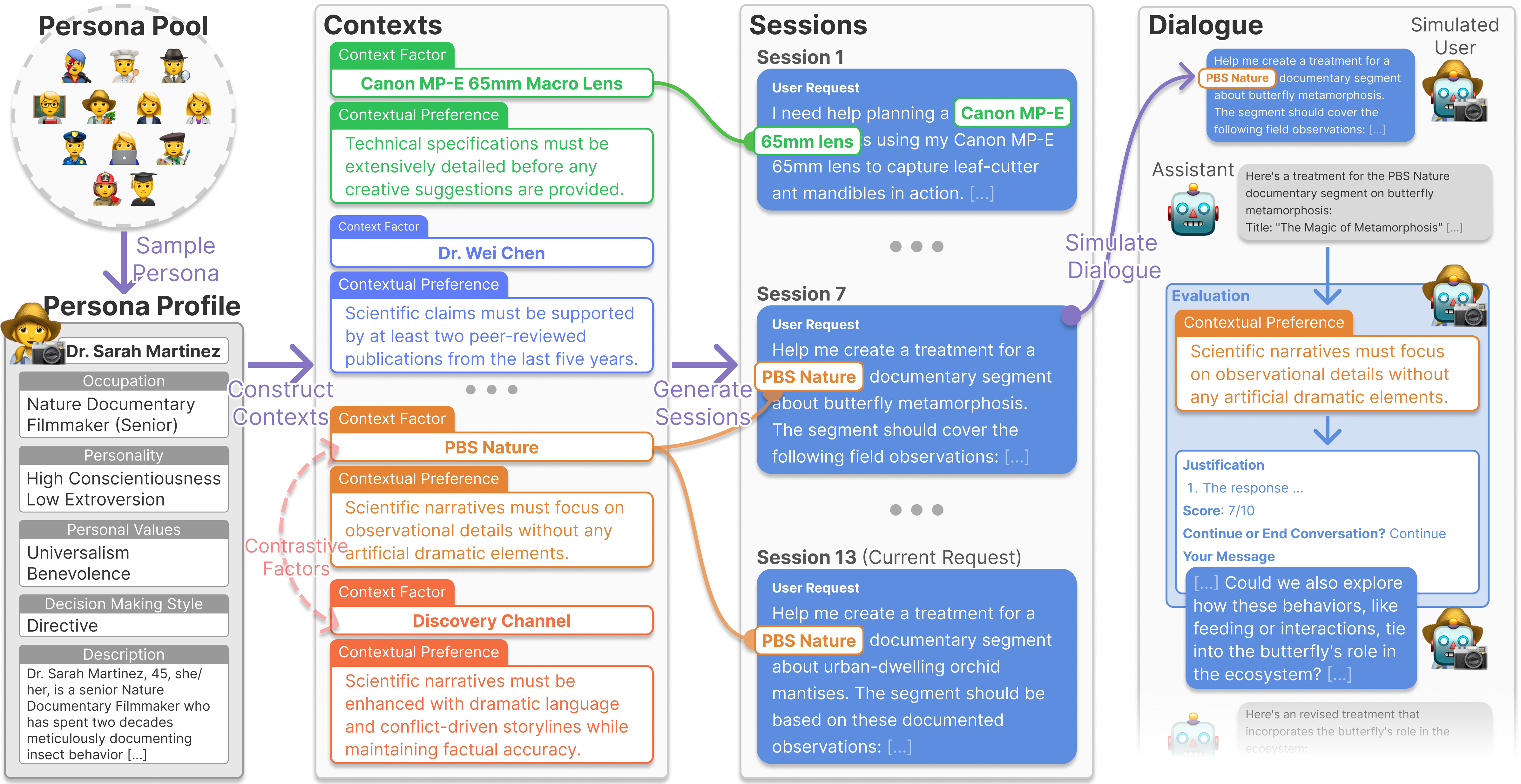}
    \vspace{-6pt}
    \caption{Data generation pipeline for \benchmarkWithEmoji{}. For each persona, we construct diverse context factors and preferences, then generate chronologically linked interaction sessions. For each session, we simulate a dialogue where the user persona evaluates an AI assistant's responses and provides feedback based on the contextual preference.}
    \label{fig:pipeline}
    \vspace{-12pt}
    \end{center} 
\end{figure*}

\vspace{-4pt}
\section{Data Generation Pipeline (\autoref{fig:pipeline})}
\vspace{-6pt}

To capture realistic interaction patterns and context-dependent user preferences~\citep{warren2011values}, we designed synthetic user interaction sessions and preferences in \benchmark{} with the following \textbf{desiderata}: 
\textbf{(1) User-specific and unique}: Focus on highly personalized contexts and preferences not inferrable from prior knowledge or commonsense; 
\textbf{(2) Indirect and gradually revealed preferences}: Preferences expressed through multiple turns of feedback and clarification to mirror real user behavior; 
\textbf{(3) Patterns of shifting preferences}: Capture how a user's preferences may shift over time or even conflict across contexts.
Unlike prior work focused on lifestyle preferences, we center our benchmark on task-oriented preferences and dialogues, reflecting real-world usage of LLMs~\citep{tamkin2024clio}.
The pipeline uses Claude 3.5 Sonnet, unless noted otherwise.
Full details in Appendix~\ref{appendix:pipeline}.

Although synthetic data may not fully represent real user behavior and diversity, we opt for synthetic data over human-collected data because it provides: (1) precise control over dataset difficulty, including how preferences are revealed and contexts are repeated; (2) higher quality by ensuring that preferences are revealed and models can infer them; and (3) enhanced diversity by varying context factor, preference, and task types.
\vspace{-8pt}

\subsection{Generation Process}
\vspace{-4pt}

\paragraph{Persona Pool}
We aimed to construct context factors and preferences that were specific and unique to a user \textbf{(Desiderata 1)}, rather than general factors (e.g., \textit{"Fender Stratocaster"} vs. \textit{"electric guitar"}) and preferences (e.g., \textit{"instructions must include precise hand positioning details"} vs. \textit{"instructions must be precise"}).
Existing persona datasets~\citep{ge2024scaling, zhou2023sotopia} lacked sufficient detail to generate unique factors and preferences.
Thus, we created a new persona pool by sampling seed persona descriptions from \citet{ge2024scaling}, combining them with attributes (e.g., personality, personal values) inspired by \citet{zhou2023sotopia}, and using an LLM to expand each into rich persona descriptions---yielding 252 distinct personas.

\vspace{-8pt}
\paragraph{Constructing Contexts}
To generate diverse yet internally coherent contexts for each persona, we instruct an LLM to first generate a list of 8 context factors and associated preferences---essentially \textit{constructing} each persona's \textit{world}.
Before each factor-preference pair, we prompt the LLM to first generate a background narrative to develop more specific and unique contexts \textbf{(Desiderata 1)}.
To increase diversity, we provide the LLM with predefined taxonomies for factor types (e.g., person, object, location, activity) and preference types (e.g., creativity, sensitivity, trustworthiness)---full lists in \autoref{appendix:types}.
To reflect how users may hold contrasting preference in different contexts \textbf{(Desiderata 3)}, the LLM also creates factor pairs $c_\text{current}$ and $c_\text{contrast}$ whose preferences conflict or contradict, $p_\text{current} = \neg p_\text{contrast}$.

\paragraph{Generating Sessions}
With each persona's list of factor-preference pairs, we prompt an LLM to generate a series of chronologically connected interaction sessions, $\textbf{S}$.
Instead of generating each session and its interactions one-by-one, we generate all the sessions first to ensure that they are logically and chronologically coherent.
When generating each session $S_i$, the LLM is instructed to first narrate a task-related situation that involves a context factor, and then generate the initial request $u_{i,1}$ where the user explicitly mentions the factor and seeks an LLM's help with the task.
To ensure that the series of sessions reveal specific contexts and preferences \textbf{(Desiderata 2)}, we prompt the LLM with a 13-session template: the final session $S_{13}$ includes $c_{\text{current}}$; two sessions $S_u$ and $S_v$ also with $c_{\text{current}}$; two sessions with $c_{\text{contrast}}$; and two sessions $S_y$ and $S_z$ ($y,z < u,v$) with $c_{\text{current}}$ but a prior preference $p_{\text{prior}} = \neg p_{\text{current}}$ that reflects a significant change in the user's preference \textbf{(Desiderata 3)}.
\vspace{-4pt}

\paragraph{Dialogue Simulation}
For each interaction session, we simulate multi-turn dialogues by using two distinct LLMs: an \textit{user simulator} role-playing as the persona and an \textit{assistant simulator} that responds to the user.
Each session begins with the user’s request \(u_{i,1}\), which the assistant answers. 
Adapting the LLM-as-a-Judge approach~\citep{zheng2023judging}, the user simulator evaluates the assistant’s response against the contextual preference \(p_i\), scoring it from 1 to 10, and then generates feedback that indirectly hints at the preference (e.g., noting the issues with the response, rather than explicitly stating the preference).
The \textit{assistant simulator} responds and the dialogue continues, gradually revealing the preference, until the \textit{user simulator} provides a score of 10 \textbf{(Desiderata 2)}.
Each dialogue is designed to contain sufficient evidence for the underlying preference to be inferrable.
\vspace{-4pt}

\paragraph{Creating Instances}
Finally, we construct instances to account for the different patterns in user preferences \textbf{(Desiderata 3)}: a user holding conflicting preferences in different contexts, and their preferences changing over time. 
Specifically, for each series of synthesized sessions, we create three types of data instances:
\vspace{-6pt}
\begin{itemize}
    \item \textbf{Consistent}: Basic instance where there is at least one $S_k\in\mathbf{S}$ such that $c_k = c_\text{current}, \quad p_k = p_\text{current} \quad \text{and} \quad u_{k,1} \neq u_\text{current}$ (i.e., at least one previous interaction session possesses the same context factor and preference as the current request). 
    \item \textbf{Contrastive}: Meets the same condition as \textit{Consistent}, and there is at least one $S_l \in \mathbf{S}$ such that $c_l \neq c_\text{current} \quad \text{and} \quad p_l = p_\text{contrast} = \neg p_\text{current}$. Here, a prior session includes a context factor with a preference that conflicts with the current context's.
    \item \textbf{Changing}: Meets the same condition as \textit{Consistent}, and there is at least one $S_m \in \mathbf{S}$ such that $m < k$, $c_m = c_\text{current}$, but $p_m = p_\text{prior} = \neg p_\text{current}$. Here, the user \textit{previously} held an opposite preference in relation to the same context factor.
\end{itemize}
\vspace{-6pt}
Each instance has 9 sessions: 8 prior interaction sessions, and the last session with $u_\text{current}$.
\vspace{-4pt}

\subsection{Data Validation}

We conducted human validation to ensure that our instances were \textbf{solvable} (i.e., preferences are inferrable from prior interactions), \textbf{challenging} (i.e., preferences not inferrable from current requests), and \textbf{realistic}.
For each instance, we validated that relevant preferences ($p_\text{current}, p_\text{contrast}, p_\text{prior}$) were fully expressed in the simulated dialogues, but $p_\text{current}$ was not expressed in  $u_\textit{current}$.
To reduce annotators' cognitive load, we employed a human-AI approach \citep{lee2022promptiverse, lee2023qasa} where an LLM first extracts user messages from the dialogues that potentially expressed each preference.
Then, two human annotators recruited via Prolific then annotated these messages and we considered that a preference was expressed if marked by at least one annotator.
Then, the authors manually revised these cases ($\sim${}9\% of all instances).
Annotators also rated the realism of the current requests at an average of 4.08 (SD=0.85) out of 5.
More details in \autoref{appendix:validation}.
\vspace{-4pt}

\subsection{Data Statistics}

\benchmark{} is composed of 756 instances, with 252 instances for each type: \textit{Consistent}, \textit{Contrastive}, and \textit{Changing}.
On average, contextual preferences are 18.4 tokens long and are decomposed into 2.90 checklist items, dialogues have 6.38 turns and are 921.5 tokens long, and prior interaction sessions are 8186.7 tokens long.
\autoref{fig:dataset-distribution} shows the distribution of types for context factors, contextual preferences, and tasks in our dataset.
\vspace{-8pt}
\section{Experiments}
\vspace{-4pt}

With \benchmarkWithEmoji{}, we evaluate open and proprietary LLMs on \textbf{inference} and \textbf{generation}.
\vspace{-4pt}

\subsection{Experimental Setup}

\paragraph{Baseline Models}
We tested a total of 10 state-of-the-art instruction-tuned LLMs and a few reasoning models: GPT-4o~\citep{hurst2024gpt}, o3-mini~\citep{openai2025o3mini}, Claude 3.7 Sonnet~\citep{anthropic2025claude37}, Claude 3.5 Sonnet~\citep{anthropic2024claude35}, Llama 3.1 405B~\citep{grattafiori2024llama}, Mistral 7B~\citep{jiang2023mistral7b}, Qwen2.5 72B~\citep{yang2024qwen2}, DeepSeek-R1~\citep{guo2025deepseek}, Gemini 2.0 Flash Thinking, and Gemini 2.0 Pro~\citep{google2024gemini2}. Details in \autoref{tab:model-details}.
\vspace{-12pt}

\paragraph{Prompts}
For the inference task, all models were zero-shot prompted (\autoref{fig:preference-infer-prompt}) to analyze prior interaction sessions and infer the preference that the user will likely hold in the current request, but is not mentioned in the request.
For the generation task, all models were zero-shot prompted (\autoref{fig:response-generate-prompt}) to generate a response for the current request by considering the possible preference that the user has based on prior interaction sessions.
\vspace{-4pt}

\paragraph{Oracle}
We test an oracle setting where models receive only prior sessions that share the current request's contextual preference, isolating the impact of identifying and retrieving relevant prior sessions.
For the Generation task, we test an oracle \textit{preference} setting where models are given the ground-truth preference when generating responses, allowing us to evaluate how their ability to adhere to preferences affects performance.
\vspace{-4pt}

\paragraph{Interaction Summary}
Recent LLM-powered AI assistants are equipped with \textit{long-term memory} allowing them to record, recall, and reason about prior user interactions~\citep{zhong2024memorybank, OpenAI2024}.
To understand how this type of intermediate representation could enable models to infer a user's preferences in different contexts, we design a setting where models are first prompted (\autoref{fig:interaction-summarize-prompt}) to summarize each dialogue in the prior interaction sessions and, then, perform the tasks with these summaries.
\vspace{-4pt}

\paragraph{Metrics (Section~\ref{sec:metrics})}
We report precision, recall, and F1-score for the \textbf{inference} task, and alignment score of model responses [1-10] for the \textbf{generation} task.
\vspace{-4pt}
\subsection{Results}

In this section, we report performance on \textbf{inference} (Section~\ref{sec:inference-results}), \textbf{generation} (Section~\ref{sec:generation-results}), and with \textbf{interaction summaries} as an intermediate representation (Section~\ref{sec:summary-results}).

\subsubsection{Inference Task}
\label{sec:inference-results}

\paragraph{Models struggle to infer contextual preferences from interactions.}
\autoref{tab:inference-full} shows the performance of all tested models on the inference task---visualized in \autoref{fig:results-visualization}.
All models struggled to adequately infer the preference relevant to the current user request from the previous interaction sessions---no model surpassed an F1 score of 60\% and with most under 50\%.
Considering how our benchmark was designed to be challenging but not overly difficult (e.g., each preference is revealed in multiple sessions, only 8 prior sessions per instance), we expect that performance will degrade significantly in more realistic settings.
\vspace{-4pt}

\paragraph{Model size and reasoning capabilities increases performance.}
Larger model size lead to better performance with the smallest model, Mistral 7B, showing the lowest performance by a notable margin of 7.8 points.
Additionally, our results show reasoning models perform the best in this task with DeepSeek-R1 and Claude 3.7 Sonnet reaching the highest performance.
This suggests that greater train-time and test-time compute enables models to better reason about user preferences and their relevant contexts from prior interactions.
As Claude 3.7 Sonnet performed best in our dataset mostly generated by Claude 3.5 Sonnet, we analyze the potential of self-enhancement bias (Appendix~\ref{appendix:self-enhancement}) but find no evidence of its presence.
\vspace{-2pt}

\begin{table}[t]
\resizebox{\textwidth}{!}{%
\begin{tabular}{@{}lccc|ccccccccc|ccc@{}}
\toprule
\textbf{} &
  \multicolumn{3}{c|}{\textbf{All Instances}} &
  \multicolumn{3}{c}{\textbf{Consistent}} &
  \multicolumn{3}{c}{\textbf{Contrastive}} &
  \multicolumn{3}{c|}{\textbf{Changing}} &
  \multicolumn{3}{c}{\textbf{Oracle Setting}} \\
\textbf{Model} &
  \textbf{P} &
  \textbf{R} &
  \textbf{F1} &
  \textbf{P} &
  \textbf{R} &
  \textbf{F1} &
  \textbf{P} &
  \textbf{R} &
  \textbf{F1} &
  \textbf{P} &
  \textbf{R} &
  \textbf{F1} &
  \textbf{P} &
  \textbf{R} &
  \textbf{F1} \\ \midrule
Mistral 7B &
  26.9 &
  30.7 &
  28.6 &
  22.3 &
  25.6 &
  23.8 &
  23.0 &
  26.7 &
  24.7 &
  35.4 &
  39.7 &
  37.4 &
  46.8 &
  59.8 &
  52.5 \\
Qwen2.5 72B &
  32.8 &
  44.9 &
  37.9 &
  30.0 &
  42.6 &
  35.2 &
  27.5 &
  36.2 &
  31.3 &
  41.0 &
  55.9 &
  47.3 &
  60.4 &
  77.1 &
  67.7 \\
Llama 3.1 405B &
  30.2 &
  45.1 &
  36.2 &
  26.4 &
  39.1 &
  31.5 &
  27.8 &
  41.3 &
  33.2 &
  36.4 &
  55.0 &
  43.8 &
  58.4 &
  77.4 &
  66.5 \\
DeepSeek-R1 &
  \underline{42.0} &
  59.5 &
  \underline{49.3} &
  \underline{44.2} &
  62.9 &
  \underline{51.9} &
  \underline{41.3} &
  \underline{58.8} &
  \underline{48.5} &
  40.6 &
  56.8 &
  47.4 &
  \underline{62.8} &
  81.2 &
  70.8 \\
GPT-4o &
  36.6 &
  52.1 &
  43.0 &
  36.5 &
  50.1 &
  42.2 &
  32.6 &
  46.4 &
  38.3 &
  40.7 &
  59.7 &
  48.4 &
  57.1 &
  77.7 &
  65.8 \\
o3-mini &
  33.6 &
  47.9 &
  39.5 &
  32.4 &
  47.9 &
  38.6 &
  31.0 &
  44.2 &
  36.5 &
  37.3 &
  51.6 &
  43.3 &
  48.9 &
  68.7 &
  57.1 \\
Claude 3.5 Sonnet &
  40.9 &
  56.9 &
  47.6 &
  40.9 &
  56.2 &
  47.4 &
  38.2 &
  52.5 &
  44.3 &
  \underline{43.6} &
  62.0 &
  51.2 &
  62.8 &
  81.6 &
  71.0 \\
Claude 3.7 Sonnet &
  \textbf{49.1} &
  \textbf{64.6} &
  \textbf{55.8} &
  \textbf{52.5} &
  \textbf{66.3} &
  \textbf{58.6} &
  \textbf{48.3} &
  \textbf{61.5} &
  \textbf{54.1} &
  \textbf{46.5} &
  \underline{65.9} &
  \textbf{54.5} &
  \textbf{69.6} &
  \textbf{84.5} &
  \textbf{76.3} \\
Gemini 2.0 Flash Thinking  & 37.8  & 53.7  & 44.4  & 37.3 & 53.0 & 43.8 & 35.5 & 49.4 & 41.3 & 40.6 & 58.6 & 48.0 & 59.5 & 76.0 & 66.8 \\
Gemini 2.0 Pro &
  40.1 &
  \underline{63.5} &
  49.2 &
  39.0 &
  \underline{63.3} &
  48.3 &
  38.3 &
  58.7 &
  46.4 &
  43.0 &
  \textbf{68.5} &
  \underline{52.9} &
  62.6 &
  \underline{83.1} &
  \underline{71.4} \\ \bottomrule
\end{tabular}
}
\caption{Precision (P), Recall (R), and F1 score for all models on the inference task in \benchmark{}, averaged across all instances, each instance type, and the oracle setting. Best results in each column are \textbf{bold}-faced and second best results are \underline{underlined}.}
\vspace{-4pt}
\label{tab:inference-full}
\end{table}

\paragraph{Retrieving relevant context can significantly increase performance.}
In the oracle setting, all models improved by approximately 20-30 points, highlighting how inference performance is strongly dependent on the capability to retrieve and focus on prior interactions with the same context.
However, even with only the relevant sessions, precision across models was still under 70\%, implying that models are still inferring less relevant details.
\vspace{-2pt}

\begin{figure}[t]
    \vspace{-12pt}
    \centering
    \begin{minipage}[t]{0.31\textwidth}
        \centering
        \includegraphics[width=\textwidth]{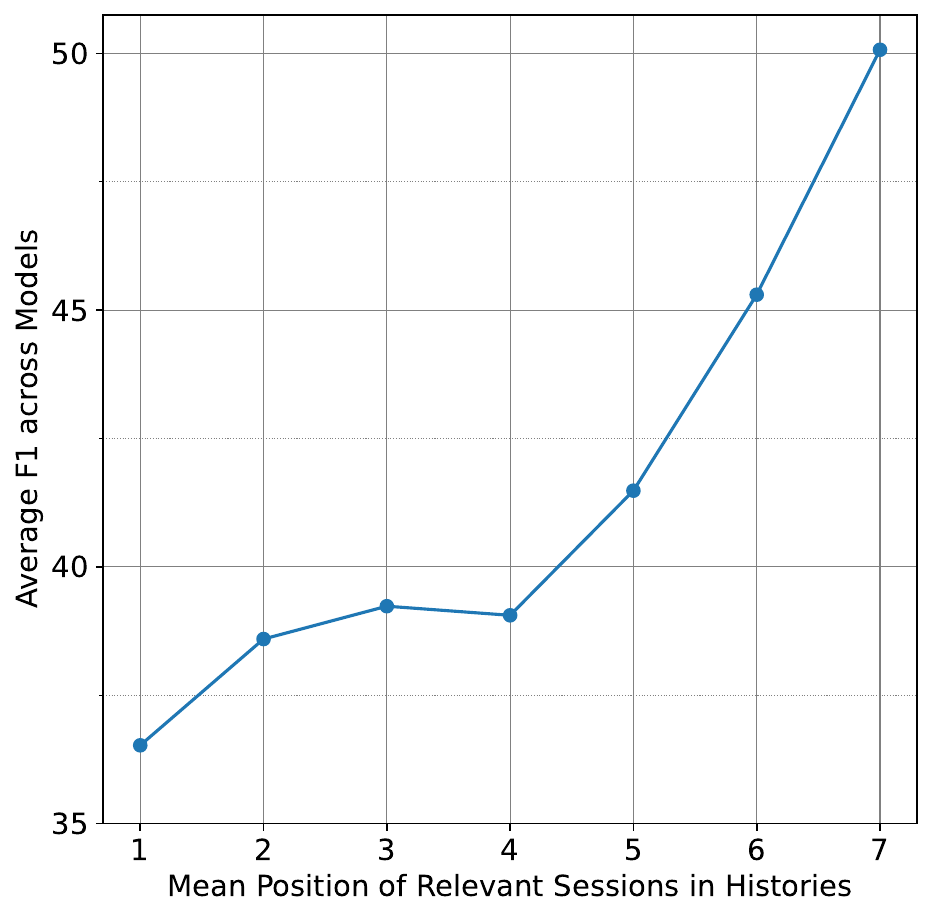}
        \caption{Mean F1 score across all models against mean position of relevant sessions in histories.}
        \label{fig:inference-position}
    \end{minipage} 
    \hfill
    \begin{minipage}[t]{0.318\textwidth}
        \centering
        \includegraphics[width=\textwidth]{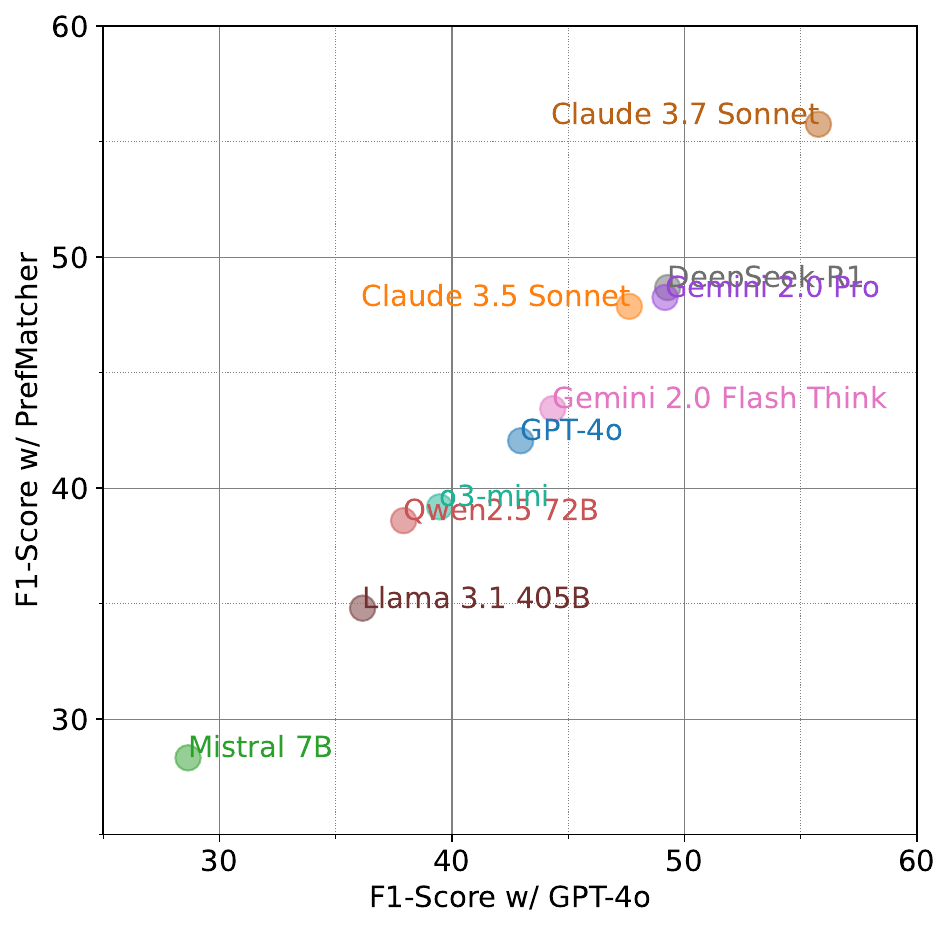}
        \caption{Correlation between F1-score when computed with GPT-4o and with \matcherName{}.}
        \label{fig:results-pref-matcher}
    \end{minipage} 
    \hfill
    \begin{minipage}[t]{0.31\textwidth}
        \centering
        \includegraphics[width=\textwidth]{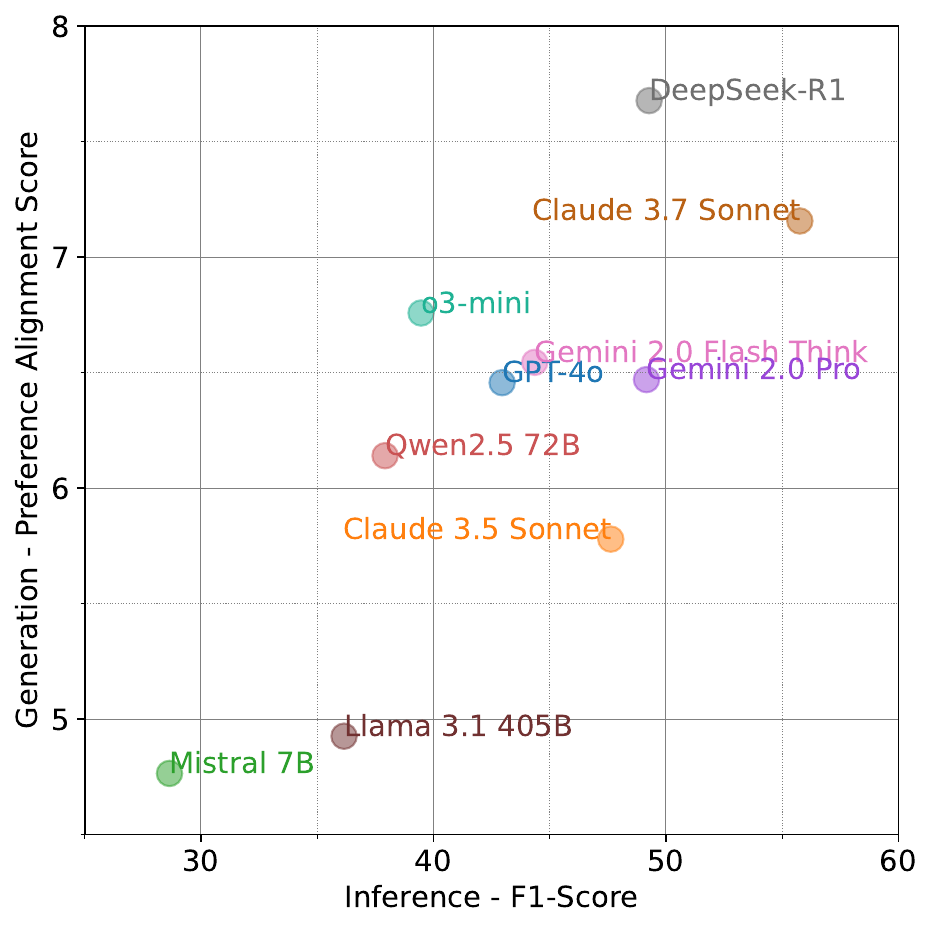}
        \caption{Inference performance against generation performance, averaged for each model.}
        \label{fig:results-correlation}
    \end{minipage} 
    \vspace{-8pt}
\end{figure}

\paragraph{Surprisingly, \textit{Changing} instances led to highest performance.}
Most models perform worse in \textit{Contrastive} instances but excel in \textit{Changing}, contrary to our expectation that models would struggle with evolving preferences. 
Deeper analysis suggests that models prioritize the \textit{most recent} sessions. 
\autoref{fig:inference-position} shows that performance improves when relevant sessions (i.e., contain the current preference) appear at the end of the history.
In \textit{Changing} instances, earlier sessions reflect the preference before a change and later ones capture the change---meaning that relevant sessions tend to be positioned towards the end.
\vspace{-2pt}

\paragraph{Error Analysis}
To understand the errors that models make when inferring preferences, we sampled and qualitatively inspected 50 responses each from DeepSeek-R1 and Llama 3.1 405B with F1 scores below 20\% ($\sim$bottom 25\%).
\autoref{tab:error_types} summarizes the identified error types.
We observe that models failed to focus on the relevant contexts, struggled to extract the specific preference from multi-turn interactions, only performed shallow inferences, or hallucinated preferences.
The models exhibited distinct error patterns: Llama 3.1 405B produced more shallow inference errors, focusing only on the current request, while DeepSeek-R1 showed less shallow errors but more incorrect context errors, reasoning over prior interactions but failing to focus on those with shared context. 
\vspace{-2pt}

{\renewcommand{\arraystretch}{1.3}
\begin{table}[b]
    \centering
    \resizebox{\textwidth}{!}{%
    \begin{tabular}{lp{9.5cm}cc}
        \toprule
        \textbf{Error Type} & \textbf{Description} & DeepSeek-R1 & Llama 3.1 450B \\
        \midrule
        Incorrect Context & The model incorrectly infers preferences from other contexts that are not relevant to the current request.& 86\% & 40\% \\
        Shallow Inference & The model infers preferences explicitly mentioned in the request or that are commonsense for the request, instead of inferring from prior interactions. & 10\% & 50\% \\
        Vagueness & The model inferred preferences that were relevant but were too broad or vague, lacking the specific details in the target preference. & 2\% & 6\% \\
        Hallucination & The model inferred preferences without clear evidence or context, likely influenced by internal assumptions or biases. & 2\% & 4\% \\
        \bottomrule
    \end{tabular}
    }
    \caption{Error types identified in preferences inferred by DeepSeek-R1 and Llama 3.1 405B, with proportion of errors that each model made for each type.}
    \label{tab:error_types}
\end{table}
}

\paragraph{Evaluation with \matcherName{} shows almost perfect correlation with GPT-4o assessments}
(\autoref{fig:results-pref-matcher}, \autoref{tab:inference-matcher}).
The Pearson correlation for model-wise average performance with GPT-4o and \matcherName{} was 0.997 (p$<$0.001).
Given this result and the strong correlation between inference and generation (discussed later), we suggest evaluating on \benchmark{} only on the inference task and with \matcherName{} to reduce cost.

\subsubsection{Generation Task}
\label{sec:generation-results}

\begin{table}[b]
\centering
\resizebox{0.85\textwidth}{!}{%
\begin{tabular}{@{}lcccccc@{}}
\toprule
\textbf{Model} &
  \multicolumn{1}{c|}{\textbf{All Instances}} &
  \textbf{Consistent} &
  \textbf{Contrastive} &
  \multicolumn{1}{c|}{\textbf{Changing}} &
  \textbf{Oracle} &
  \textbf{Preference} \\ \midrule
Mistral 7B & \multicolumn{1}{c|}{4.76} & 4.03 & 4.20 & \multicolumn{1}{c|}{6.06} & 6.12 & 7.36          \\
Qwen2.5 72B     & \multicolumn{1}{c|}{6.14} & 5.60 & 5.38 & \multicolumn{1}{c|}{7.45} & 7.33 & 8.69          \\
Llama 3.1 405B  & \multicolumn{1}{c|}{4.93} & 4.43 & 4.29 & \multicolumn{1}{c|}{6.06} & 6.96 & 8.84          \\
DeepSeek-R1 &
  \multicolumn{1}{c|}{\textbf{7.68}} &
  \textbf{7.50} &
  \textbf{7.33} &
  \multicolumn{1}{c|}{\textbf{8.21}} &
  \textbf{8.95} &
  \underline{9.66} \\
GPT-4o                   & \multicolumn{1}{c|}{6.46} & 5.88 & 5.75 & \multicolumn{1}{c|}{7.74} & 7.79 & 9.20          \\
o3-mini                  & \multicolumn{1}{c|}{6.76} & 6.31 & 6.04 & \multicolumn{1}{c|}{\underline{7.93}} & 8.10 & \textbf{9.72} \\
Claude 3.5 Sonnet        & \multicolumn{1}{c|}{5.78} & 5.25 & 5.08 & \multicolumn{1}{c|}{7.01} & 8.05 & 9.41          \\
Claude 3.7 Sonnet        & \multicolumn{1}{c|}{\underline{7.16}} & \underline{7.13} & \underline{6.50} & \multicolumn{1}{c|}{7.84} & \underline{8.61} & 9.63          \\
Gemini 2.0 Flash Thinking & \multicolumn{1}{c|}{6.54} & 6.03 & 5.91 & \multicolumn{1}{c|}{7.69} & 7.58 & 9.43 \\
Gemini 2.0 Pro           & \multicolumn{1}{c|}{6.47} & 6.29 & 5.81 & \multicolumn{1}{c|}{7.31} & 7.81 & 9.45          \\ \bottomrule
\end{tabular}
}
\caption{Preference alignment scores for all models on the generation task in \benchmark{}, measured for all instance types, each instance type, oracle setting, and oracle preference setting.}
\label{tab:generation-full}
\end{table}

\paragraph{Performance in the generation task follows similar trends to the inference task.} 
\autoref{tab:generation-full} shows the models' performance in the generation task---visualized in \autoref{fig:results-visualization}.
Similar to the inference task, we observe that model performance is generally low, increases with model size and reasoning capabilities, increases significantly in the oracle setting, and is highest in Changing instances while being generally lower in Contrastive.
\vspace{-4pt}

\paragraph{Strong correlation between the inference and generation tasks.} 
\autoref{fig:results-correlation} shows a positive correlation between model performance in the inference and generation tasks.
The Pearson correlation for model-wise average performance in the tasks was 0.764 (p=0.010), suggesting that the inference task can serve as a proxy to evaluate models' ability to generate contextually personalized responses.
Interestingly, certain models excelled at one task over the other: Claude 3.7 Sonnet led in inference, but was outperformed by DeepSeek-R1 in generation.
\vspace{-4pt}

\paragraph{Generation performance is not tied to response generation capabilities.}
Almost all of the models show significantly high performance in the oracle preference setting (i.e., generating a response given the ground-truth preference).
This indicates that the low performance across the models is not due to the models' inability to generate responses that satisfy the preference, but rather due to their inability to precisely infer the relevant preference.
\vspace{-4pt}

\subsubsection{With Summaries}
\label{sec:summary-results}
\vspace{-4pt}

\begin{figure}[t]
  \vspace{-8pt}
  \centering
  \begin{minipage}{0.495\textwidth}
    \centering
    \includegraphics[width=\textwidth]{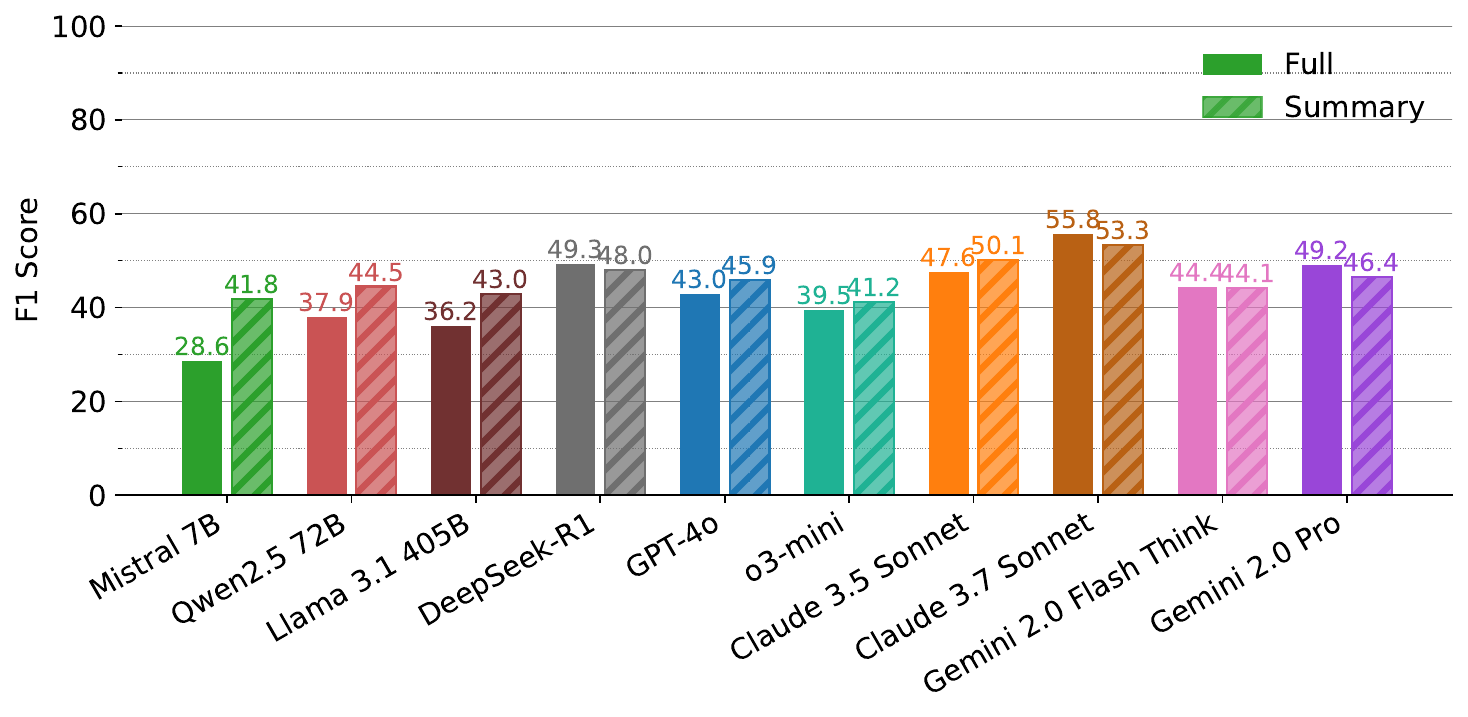}
  \end{minipage}
  \hfill
  \begin{minipage}{0.495\textwidth}
    \centering
    \includegraphics[width=\textwidth]{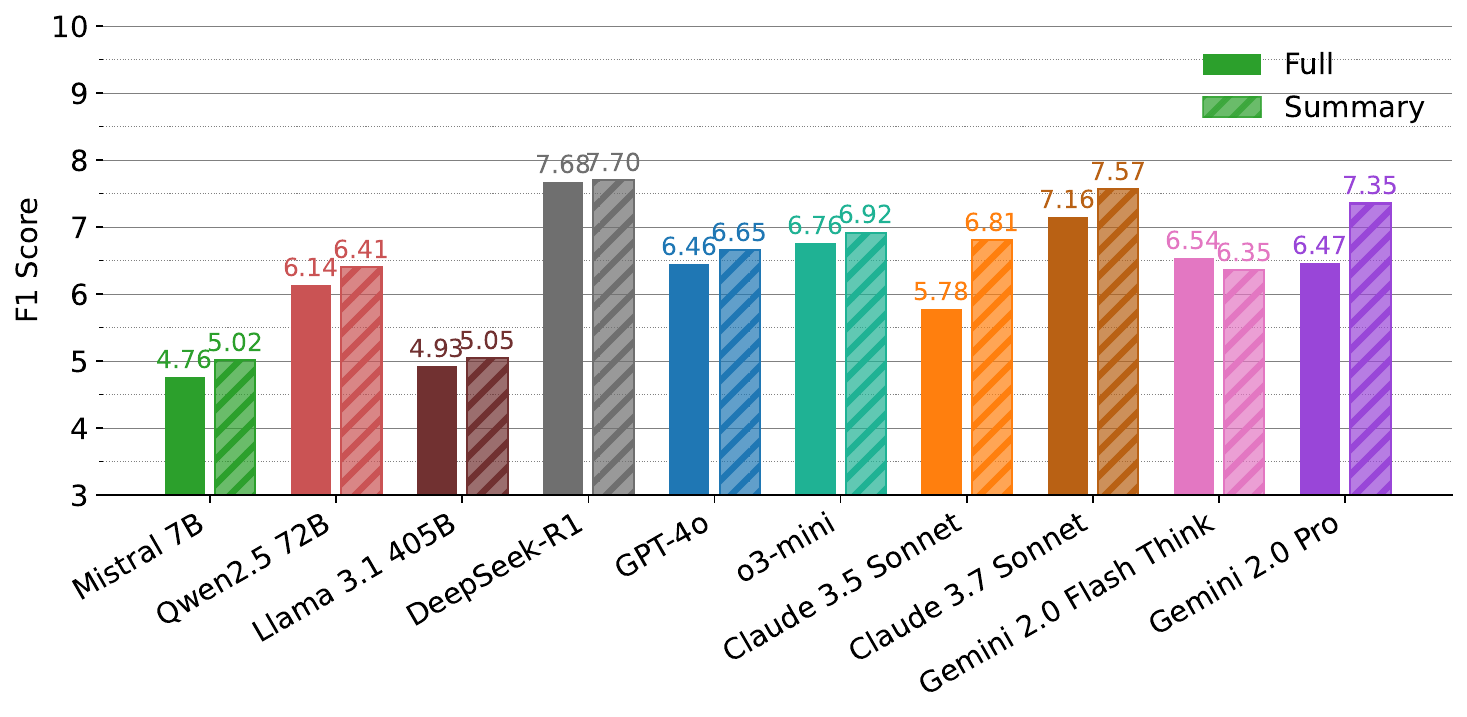}
  \end{minipage}
  \vspace{-8pt}
  \caption{Comparison of each model's results for the inference task (left) and generation task (right) with the full prior interaction sessions or summaries of these sessions.}
  \vspace{-8pt}
  \label{fig:results-summary}
\end{figure}

\paragraph{Summaries have an \textit{equalizing} effect across models.}
\autoref{fig:results-summary} compares model performance with and without summaries---full results in \autoref{tab:inference-summary} and \autoref{tab:generation-summary}.
Summaries offer minimal gains for strong models---slightly lowering performance for reasoning models---yet substantially boosts weaker models, bringing them closer to the performance of strong models.
This suggests that summaries help weaker models to reason about and extract preferences from each session, but may lead to information loss for strong models.
Notably, the smallest model, Mistral 7B, shows a substantial increase of 13 points in inference, suggesting potential for local and privacy-preserving LLM personalization.
\vspace{-8pt}

\paragraph{Even with summaries, models show low precision.}
We observe that the increase in inference performance with the summaries is mostly attributed to recall.
This indicates that the summaries are not necessarily helping the models focus on the relevant interaction sessions, but rather it is helping models extract the preferences from all the sessions.

\subsubsection{Practical Implications}

Our findings suggest three actionable directions for personalized LLMs.
First, integrate retrieval techniques that identify prior sessions from users' interaction histories that are \textit{contextually} relevant to the current request, as oracle results show 20-30 point improvements.
Second, when deploying smaller or local LLMs, cache summaries of each interaction session focusing on context and preferences. 
Finally, prompt or tune models to perform reasoning about users' underlying preferences during multi-turn interactions, rather than simply inferring from surface-level expressions.

\vspace{-4pt}
\section{Related Work}
\vspace{-4pt}

\paragraph{Personalized LLMs}

To reduce harms and increase helpfulness, research has explored how to align LLMs with human values~\citep{bai2022constitutional, bai2022training} by training them on \textit{general} preferences~\citep{bai2022constitutional, bai2022training, ouyang2022training, cui2023ultrafeedback, kim2023aligning}. 
Recent work has taken a more personalized view to alignment, considering individual preferences and values~\citep{sorensen2024roadmap, tseng2024two}.
For example, \citet{kirk2024prism} collected a preference dataset with detailed user information (e.g., demographics, behavior attributes). 
Other work explored prompting LLMs with user profiles~\citep{yang2023palr, wang2024ai}, interaction logs~\citep{baek2024knowledge}, or user-written artifacts~\citep{salemi2023lamp, salemi2024optimization}. 
Alternatively, \citet{jang2023personalized} and \citet{lee2024aligning} fine-tuned LLMs on decomposed preferences to produce diverse responses for users.
More recently, \textsc{PrefEval}~\citep{zhao2025llms} evaluates whether LLMs can identify user's global preferences from long-context conversations.
In this work, we take a more nuanced view to LLM alignment by considering that each individual has distinct preference in different contexts.
\vspace{-4pt}

\paragraph{Interactions with LLMs}

LLMs possess the capability to interact with human users~\citep{wang2023interactive}.
User-LLM interactions can organically reveal details about user intents and preferences~\citep{kim2024understanding, shi2024wildfeedback}, enabling models to align themselves with users.
Recent work explores how to extract feedback from user-LLM interactions~\citep{don2024learning, wu2024aligning}, and how to enhance LLMs' \textit{memory} to leverage user knowledge from interaction histories~\citep{zhong2024memorybank, wang2024augmenting, zhong2022training, wu2024longmemeval, kim2024dialsim}.
Other work focused on training LLMs to capture richer details from user interactions by clarifying intents or eliciting information~\citep{qian2024tell, li2023eliciting, andukuri2024star}.
In this work, we evaluate LLMs' capabilities to infer the preferences that users hold in different contexts from interaction histories.



\vspace{-8pt}
\section{Conclusion}

In this work, we introduce \benchmarkWithEmoji{}, a challenging benchmark that assesses LLMs' ability to infer user's contextual preferences from interaction sessions and apply these in future contexts.
Our benchmark consists of task-oriented dialogues where users gradually and indirectly reveal their preferences through feedback, assesses performance in inference and generation tasks, and employs LLM-based evaluation methods---including a fine-tuned LLM metric.
Experiments with 10 state-of-the-art LLMs demonstrate that LLMs struggle to both extract users' preferences from interactions and to discern which preferences are relevant to which contexts---precision under 50\% and recall under 65\% for all models.
While explicit memory (i.e., interaction session summaries) benefits smaller models, they have minimal or even negative impact on larger models---underscoring the need for further improvements.
\benchmark{} highlights LLMs' current limitations in supporting personalized interactions, and serves as a valuable resource for advancing conversational AI systems for interactive and contextual personalization.
In practice, we recommend evaluating models on the inference task of \benchmark{} by using \matcherName{}.
\section{Ethics Statement}

In this paper, we introduce \benchmark{}, a benchmark designed to evaluate LLMs' ability to infer user's contextual preferences from interaction sessions.
Since the data in our benchmark is newly created, we conducted a rigorous human validation process. 
Specifically, during the human validation that assessed that whether the data instances were solvable, human annotators were also asked to flag any possible data points that appeared harmful or offensive.
Our human data validation protocol has been determined exempt by the IRB of the author's institution.
Additionally, the authors manually inspected all of the data points (i.e., context factors and contextual preferences) to identify and filter out any that appeared offensive, harmful, or unethical.
We conducted all experiments using either publicly available models or through documented commercial API access. 
We have detailed the API versions and configuration of all of the evaluated LLMs for reproducibility.
We acknowledge that LLMs may inherit biases from their training data, potentially leading to our dataset incorporating these same biases and not being fully reflective of real users.
To mitigate this issue, we aimed to increase the diversity of the user personas, context factors, contextual preferences, and conversation topics. Namely, we created taxonomies for each of these components to guide the generations to reflect diverse types for each of these components. 
Finally, we acknowledge personalized conversational AI systems can raise concerns related to user's privacy as user information is stored, recalled, and analyzed.
It is crucial that such AI systems implement mechanisms to anonymize data, provide mechanisms for users to control what data is stored, or support local, on-device storing and processing to ensure that user information is never transmitted externally.
To promote reproducibility and advance research in this field, we have made our benchmark dataset, code, and model publicly available.
\section*{Acknowledgments}

This work was supported by NAVER-KAIST Hypercreative AI Center and by Institute of Information \& communications Technology Planning \& Evaluation (IITP) grant funded by the Korea government(MSIT) (No. RS-2024-00443251, Accurate and Safe Multimodal, Multilingual Personalized AI Tutors).
We would like to express our gratitude to Hyunwoo Kim for his invaluable discussion and insights.
We would also like to thank Yuewen Yang for their help during early stages of this project. 
We also thank all of our Prolific participants that helped us validate and construct our dataset.

\bibliography{references}
\bibliographystyle{colm2025_conference}

\appendix

\section{Metrics}

\subsection{Preference Match}
\label{appendix:metrics-match}

To measure the similarity between inferred and ground-truth preferences, we use two LLM prompts: preference decomposer, which decomposes preferences into multiple checklist questions (\autoref{fig:preference-decompose-prompt}), and preference-checklist matcher, which assesses how much each question in a checklist matches a preference (\autoref{fig:preference-match-prompt}). 
In our experiments, we use \texttt{gpt-4o-2024-11-20} through the OpenAI API with a temperature of 0 for both steps.

\paragraph{Meta-Evaluation for Preference Match}

\begin{figure}[h]
    \centering
    \includegraphics[width=\textwidth,height=0.98\textheight,keepaspectratio]{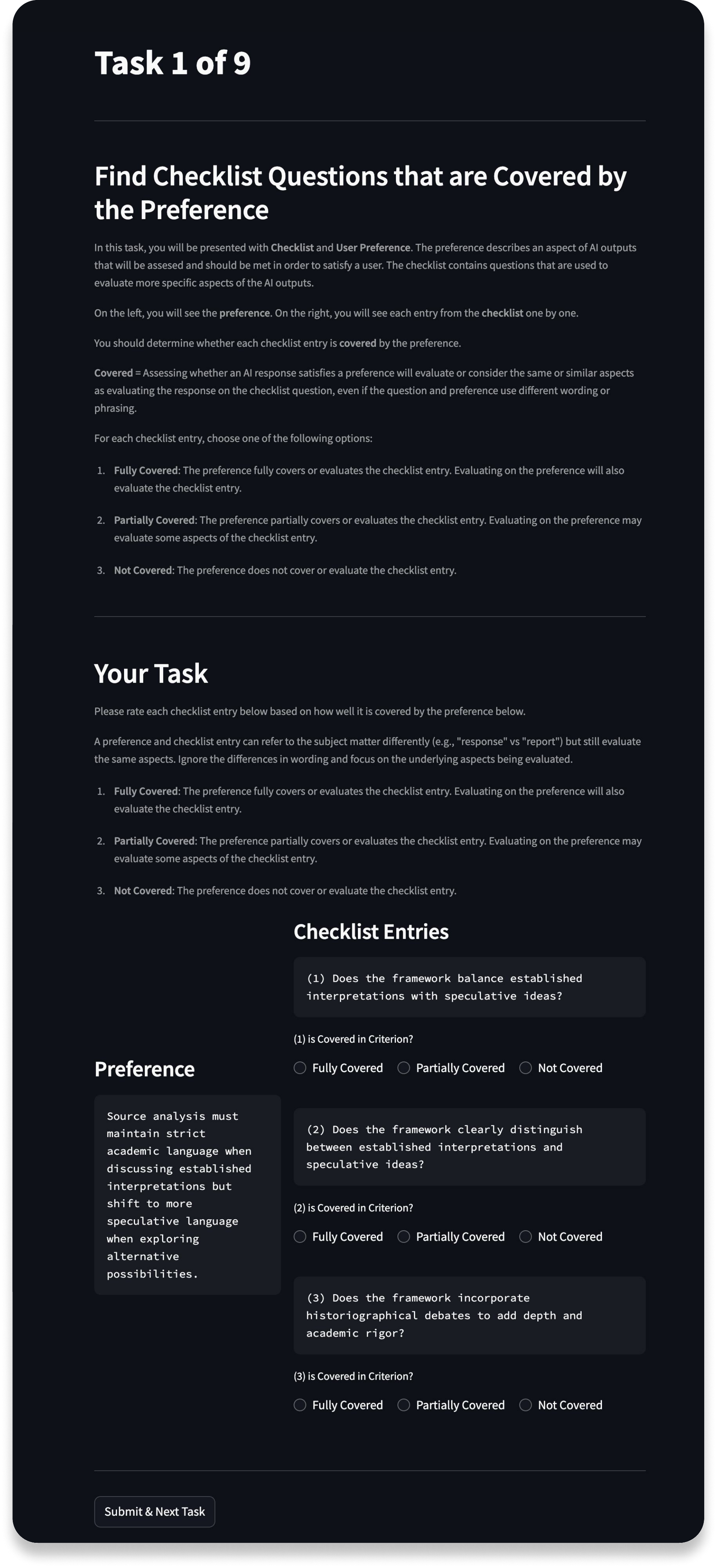}
    \caption{Annotation interface for creating the dataset to evaluate the performance of models in preference matching.}
    \label{fig:match-interface}
\end{figure}

Using our pipeline, we created an additional 240 preference-checklist pairs.
Then, we recruited annotators through Prolific to rate the match between preference and the entries in the checklist for each pair.
In total, we recruited 90 annotators, where all passed a qualification task, and each annotator inspected a total of 8 preference-checklist pairs.
Annotators spent an average of around 10 minutes in the task and were compensated £2.25 for their time.
In our interface (\autoref{fig:match-interface}), annotators were presented with a preference and each item from a checklist side-by-side, and were asked to rate whether each item was "Fully Covered", "Partially Covered" or "Not Covered" by the preference.
Through this, we obtained 3 annotations per preference-checklist pairs.
The agreement between annotators resulted in a Krippendorff's alpha of 0.467 (\textit{moderate agreement}).
Due to the moderate agreement, for each checklist item in a pair, we found the rating (i.e., "not covered", "partially covered", or "fully covered") with the majority vote and this was set as the rating for that checklist item.
We removed any checklist items without a majority vote, where 2 checklist-preference pairs were removed entirely as all the checklist items had no majority vote.
This resulted in 238 preference-checklist pairs, where 8 pairs were used as few-shot examples in the prompt and the remaining 230 pairs were used for the meta-evaluation. 
Each checklist in a pair had an average of 3.31 checklist items, leading to a total of 762 match ratings.

\vspace{-4pt}
\paragraph{Finetuning Details for \matcherName{}}
To finetune \matcherName{}, we first synthesized a separate dataset of 64 additional user personas and their interaction sessions with our pipeline.
For each persona, we created 10 data instances: Consistent, Contrastive, Changing, and separate instances for each session that does not include the current preference in the other three instances.
We then selected three models that reflected a range of performance in the inference task (i.e., Mistral-7B-Instruct-v0.3, Qwen2.5-72B-Instruct, and DeepSeek-R1) and evaluated them on the created instances using our pipeline.
This resulted in 2 matched preference-checklist pairs for each model and instance (i.e., predicted preference matched with the ground-truth checklist, and ground-truth preference matched with the predicted checklist)---total 6 per instance and a total of 4,224 samples, which we use for finetuning.
We take Qwen2.5-7b-Instruct as the base model and finetune it through QLoRA with the hyperparameters reported in \autoref{tab:lora_hyperparameters}.

\begin{wraptable}{r}{0.4\textwidth} %
    \vspace{-24pt}
    \centering
    \resizebox{0.4\textwidth}{!}{
    \begin{tabular}{ll}
        \toprule
        \textbf{Hyperparameter} & \textbf{Value} \\
        \midrule
        Epochs & 1 \\
        Per-device train batch size & 4 \\
        Gradient accumulation steps & 8 \\
        Learning rate & $3 \times 10^{-4}$ \\
        Weight decay & $1 \times 10^{-2}$ \\
        Optimizer & AdamW (torch) \\
        LR scheduler type & Cosine with warmup \\
        Number of warmup steps & 100 \\
        LoRA Rank ($r$) & 64 \\
        LoRA Alpha & 128 \\
        LoRA Dropout & 0.0 \\
        LoRA Target Modules & Q proj, V proj, Output proj \\
        Apply LoRA to MLP & True \\
        \bottomrule
    \end{tabular}
    }
    \caption{Summary of LoRA fine-tuning hyperparameters.}
    \label{tab:lora_hyperparameters}
    \vspace{-10pt}
\end{wraptable}

\vspace{-4pt}

\subsection{Preference Alignment}
\label{appendix:metrics-quality}

To evaluate the alignment of generated responses with the ground-truth preference, we use the LLM-as-a-Judge method~\citep{zheng2023judging} with \texttt{gpt-4o-2024-11-20} and temperature of 0. 
When evaluating, the LLM receives the user's request, the model's response, the ground-truth preference, and the checklist that the preference was decomposed into.
We adapt the prompt from \citet{cook2024ticking} that evaluates LLM outputs with checklists, which demonstrated higher consistency with human preferences.
Full prompt in \autoref{fig:response-judge-prompt}.

\vspace{-4pt}
\section{Supplementary Details for Data Generation Pipeline}
\label{appendix:pipeline}

\vspace{-4pt}
We mostly employed \texttt{claude-3-5-sonnet-20241022} through the Amazon Bedrock API.

\begin{wraptable}{r}{0.4\textwidth} %
    \vspace{-4pt}
    \centering
    \resizebox{0.4\textwidth}{!}{
\centering
\begin{tabular}{ll}
\toprule
\textbf{Category} & \textbf{Possible Values} \\
\midrule
Career & Entry-level/Beginner/Novice \\
Level  & Mid-level/Intermediate/Associate \\
             & Senior-level/Advanced/Expert \\[0.5em]
Personality & Openness to Experience (High/Low) \\
Traits & Conscientiousness (High/Low) \\
             & Agreeableness (High/Low) \\
             & Neuroticism (High/Low) \\[0.5em]
Personal & Self-Direction \\
Values & Achievement \\
             & Universalism \\
             & Power \\
             & Stimulation \\
             & Benevolence \\
             & Security \\
             & Tradition \\
             & Conformity \\
             & Hedonism \\[0.5em]
Decision- & Analytical \\
Making & Directive \\
Style   & Conceptual \\
             & Behavioral \\
\bottomrule
\end{tabular}
}
\caption{User persona profile trait categories and their possible values.}
\label{tab:template_traits}
\end{wraptable}

\vspace{-4pt}
\subsection{Generating User Personas}

To generate $k$ personas, our pipeline first constructs $k$ persona templates consisting of the following components: a seed persona description, a career level, personality traits, personal values, and a decision-making style.
Each template is constructed by balanced random sampling of values for each of component.
For the seed descriptions, we sample persona descriptions from PersonaHub~\citep{ge2024scaling}, where each description is a sentence that describes the persona's occupation, and their responsibilities or interests.
We noticed that PersonaHub's personas skewed heavily towards a subset of occupations.
To diversify our persona pool, we first collected all the tokens from the persona descriptions in the dataset and sorted them based on frequency. 
Then, we manually selected tokens that represented distinct occupations, which resulted in 115 highly-common occupation tokens.
To sample seed descriptions, we first sample $k$ distinct occupation tokens from this list, and then select random descriptions that contain each sampled token.

Inspired by \citet{zhou2023sotopia}, we expand each template with additional persona characteristics and behavioral traits.
Specifically, each template includes attributes across four categories: a career level, two traits from the Big Five personality traits~\citep{goldberg1992development}, two values from the Schwartz's basic human values~\citep{cieciuch2012comparison}, and a decision-making style~\citep{hamilton2016development}. 
The detailed traits and their possible values are summarized in \autoref{tab:template_traits}. 
For a set of templates, we prompt \texttt{claude-3-5-sonnet-20241022} (\autoref{fig:profile-generate-prompt}) with temperature of 0.7 to generate a 6-sentence persona description for each template, which enriches the persona with a name and a narrative of their needs, challenges, background, typical behaviors, etc.

\subsection{Generating Context Factors}

Given a persona description, our pipeline generates a list of 8 context factors and associated preferences for that persona by prompting \texttt{claude-3-5-sonnet-20241022} (\autoref{fig:context-factors-prompt}) with temperature of 0.7.
All of these factors are generated at once to guide the LLM to generate a list of mutually consistent and coherent context factors.
To generate each factor, we instruct the LLM to first generate a background story in the persona's life to elaborate on the persona's broader world and guide it to generate more unique and specific factors.
Then, the LLM selects a type for the factor from a pre-defined list and a set of task types that could be influenced by the factor---lists for these types in Appendix~\ref{appendix:types}.
After naming the context factor, the LLM selects a preference type and generates a preference relevant to this context factor and preference type.
While we could have guaranteed an equal distribution for factor, task, and preference types by sampling and pre-selecting these for the LLM, we observed that allowing the LLM to select these types led to more coherent factors and preferences, and increased the coherency between them.
In the same prompt, the LLM is also instructed to ensure that the last two context factors are contrastive: the context factors are of the same type and similar, but their contextual preferences should contradict with each other and be mutually exclusive.

\subsection{Generating Interaction Sessions}

Given a persona description and a list of context factor-preference pairs, our pipeline generates a series of 13 unique interaction sessions by prompting \texttt{claude-3-5-sonnet-20241022} (\autoref{fig:interaction-sessions-prompt}) with temperature of 0.7.
Specifically, the LLM is prompted to generate a series of chronologically connected situations where the user is performing a specific task and decides to seek the help of the AI assistant.
For each scenario or session, the LLM selects a context factor and its preference from the given list, selects a task type that the user will perform in the session, generates a story that describes the situation, and then generates the request that the user would ask to the AI assistant in this situation.
We noticed that the LLM frequently mentioned or implied the preference in the user's request, which would reveal the preferences without requiring inference from the interactions.
To avoid mentioning the preference in the request, the LLM was instructed to first generate a version of the request that includes the preference and then generate a version without it.
Considering that real-world users frequently provided additional context or resources in their requests when performing tasks (e.g., documents, code, data), we also instruct the AI assistant to separately generate these resources if necessary in a session.

To construct the various types of instances in our dataset (i.e., Consistent, Contrastive, and Changing), the LLM is provided with a template for the sequence of interaction sessions that specifies which context factors should be included in which sessions.
Specifically, we first randomly select one factor from the contrastive pair to be the \textit{current factor}, while the other factor in the pair is considered to be the \textit{contrastive factor}.
Then, the template is constructed by first randomly selecting 4 positions within the sequence (excluding the final position) and specifying that each of these sessions should include the \textit{current factor}---ensuring that the \textit{current factor} appears in random positions in the series of sessions.
Among these 4 positions with the \textit{current factor}, the template also specifies that the first two positions should include the original preference related to the \textit{current factor}, but the last two should share a new contextual preference, \textit{changed preference}, that contrasts or conflicts with the original preference. 
This reflects a significant change in the preference related to the \textit{current factor}.
From the remaining positions but excluding the final position, two positions are randomly selected and the template specifies that these should include the \textit{contrastive factor} and its related preference.
Finally, the template specifies that the last position should include the \textit{current factor}.

\subsection{Simulating User-LLM Dialogues}

For each interaction session, our pipeline then creates user-LLM dialogue by simulating a dialogue between an LLM simulating the user persona and an LLM simulating an AI assistant. 
For the user simulator, we employ \texttt{gpt-4o-2024-11-20} with temperature of 0.3 as the user simulator (prompt in \autoref{fig:user-simulation-prompt}).
At each turn in the dialogue, the user simulator is provided with the persona's profile description, the context factor and contextual preference for that interaction session, the checklist for that preference, and all the messages up to that turn between the user simulator and the assistant simulator.
The user simulator is prompted to evaluate the assistant's most recent response on each of the checklist items and to provide a score, ranging from 1 to 10, to the response.
To ensure that each message only partially implies the preference, we instruct the user simulator to select a subset of checklist items to reference or imply in their message.
Based on this selection, the user simulator is then prompted to generate its message to the assistant simulator while ensuring that its concise and indirect (e.g., instead of stating the preference, indicating part of the assistant's response that misaligned with the preference).
When the evaluation score is a 10, the user simulator is instructed to end the conversation but it is still instructed to generate a final positive message to the AI assistant. 
While this may differ from realistic user-LLM dialogues where the LLM's message is always the last, we observed that this was a necessary to ensure that all aspects of a preference were mentioned at least once in each dialogue.

For the assistant simulator, we employ \texttt{Llama-3.1-8B-Instruct-Turbo} with temperature of 0.7.
We used a smaller and relatively weaker model for the assistant simulator as we expected that this model would require more turns to fully satisfy the user simulator's feedback---leading to more opportunities for the user simulator to implicitly reveal the relevant contextual preference.
For the assistant simulator, we only provide a simple system prompt that instructs it to generate relatively concise responses to ensure that the overall dialogues are not excessively lengthy: "\texttt{You are a conversational assistant. Limit your answers to 6 sentences.}"

\subsection{Creating the Instances}

Our dataset consists of three types of instances: \textit{Consistent}, \textit{Contrastive}, and \textit{Changing}.
For each persona, we generate 13 interaction sessions and we construct these instances by removing specific sessions from this list. Specifically:
\begin{itemize}
    \item \textit{Consistent}: To create the Consistent instance, we remove the two interaction sessions that include the \textit{current factor} but with the \textit{changed preference}. We additionally remove the two interaction sessions that include the \textit{contrastive factor}. As a result, the instance contains 8 prior interaction sessions, where two include the \textit{current factor} with its original preference, and the final session that also includes the \textit{current factor}.
    \item \textit{Contrastive}: To create the Contrastive instance, we again remove the two interaction sessions that include the \textit{current factor} but with the \textit{changed preference}. We additionally select and remove two random sessions that do not include the \textit{current factor} or the \textit{contrastive factor}.  As a result, the instance contains 8 prior interaction sessions, where two include the \textit{current factor} with its original preference and two include the \textit{contrastive factor}, and the final session that also includes the \textit{current factor}.
    \item \textit{Changing}: To create the Changing instance, we remove the two interaction sessions that include the \textit{contrastive factor}. We additionally select and remove two random sessions that do not include the \textit{current factor}.  As a result, the instance contains 8 prior interaction sessions, where two include the \textit{current factor} with its original preference and two include the \textit{current factor} but with the \texttt{changed preference}, and the final session that also includes the \textit{current factor}.
\end{itemize}

\section{Data Examples}
\label{appendix:data-examples}

\autoref{tab:dataset-examples} presents a couple of example user persona profiles, context factors, and associated contextual preferences from our dataset.

\renewcommand{\arraystretch}{1.25}
\begin{table}[h]
\centering
\resizebox{\textwidth}{!}{%
\begin{tabular}{p{6cm}p{3cm}p{8cm}}
\toprule
\textbf{Persona Descriptions} &
  \textbf{Context Factors} &
  \textbf{Contextual Preferences} \\ \midrule
\multirow{3}{=}{Sarah Chen, 24, she/her, recently started pursuing amateur astronomy while working as a junior accountant, spending her evenings studying the night sky. She yearns to master the traditional constellations and their mythologies, seeking validation from more experienced astronomers in her local club, though she often doubts {[}...{]}} &
  Celestron NexStar 6SE Telescope &
  Instructions must include explicit warnings about what actions to avoid, with detailed explanations of potential consequences. \\
  \cline{2-3}
 &
  Astrophotography Sessions &
    Each step in the process must be broken down into micro-steps with specific validation checkpoints throughout. \\
  \cline{2-3}
 &
  Family Stargazing Night &
  Explanations must incorporate narrative elements and personal connections while maintaining scientific accuracy. \\ \midrule
\multirow{3}{=}{Marco Santos, 24, he/him, is an entry-level cartographer at a geographical research institute in Barcelona, bringing fresh energy to traditional mapping techniques. He seeks to make his mark by creating innovative visualization methods for Catalonia's diverse landscapes and dreams of developing an {[}...{]}} &
  Dr. Elena Martí &
  Every proposed innovation must be justified with at least three established cartographic principles from standard textbooks. \\
  \cline{2-3}
 &
  Catalonian Environmental Agency &
  All documentation must follow a structured format with numbered sections and subsections, maintaining formal technical language throughout. \\
  \cline{2-3}
 &
  Barcelona Tech Hub &
  Documentation should use bullet points and informal language, focusing on quick capture of essential information only. \\ \midrule
\multirow{3}{=}{Marcus Chen, 34, he/him, is a high school chemistry teacher who brings an energetic presence to his classroom. He strives to make science accessible and exciting for his students by drawing unexpected parallels between sports and chemical reactions, constantly seeking ways to help athletes in the school's teams understand how science impacts {[}...{]}} &
  Chemistry Lab Storage Room &
  Storage and inventory documentation must specify exact shelf locations and quantities with zero ambiguity or room for interpretation. \\
  \cline{2-3}
 &
  Regular Chemistry Class Slides &
  Slides must prioritize visual demonstrations and animations while including hidden detailed text notes that can be revealed when needed for absent students. \\
  \cline{2-3}
 &
  AP Chemistry Class Slides &
  Slides must prioritize comprehensive written explanations while relegating visual elements to small supporting diagrams in corners. \\ \midrule
\multirow{3}{=}{Marcus Thompson, 32, he/him, is an established freelance photographer specializing in landscape and cultural photography. He thrives on chasing the perfect shot, often embarking on spontaneous road trips and scaling challenging terrains to capture unique perspectives that will stand out in the competitive photography market {[}...{]}} &
  Instagram Photography Community &
  Content should emphasize personal creative vision while acknowledging audience preferences as secondary considerations. \\
  \cline{2-3}
 &
  Adobe Lightroom &
  Software guidance should focus on workarounds and customizations that prioritize creative freedom over organizational efficiency. \\
  \cline{2-3}
 &
  Capture One Pro &
  Software guidance should emphasize mastering complex features that enable precise control over the creative process. \\ \midrule

\multirow{3}{=}{Richard Blackwood, 58, he/him, is a senior-level philatelist who curates one of the largest private collections of military-themed stamps in Europe. His meticulous approach to organizing and documenting his collection has earned him recognition in philatelic societies, where he regularly presents his research on rare military {[}...{]}} &
Auction Bidding Protocol &
Bidding analysis must establish three distinct price thresholds with specific justification for each based on historical auction data. \\
  \cline{2-3}
 &
Monthly Philatelic Society Presentations &
Complex technical concepts must be introduced through a three-step process: historical context, visual example, and technical explanation. \\
  \cline{2-3}
 &
Microscopic Authentication Process &
Authentication must document every visible detail regardless of time required, creating comprehensive photographic evidence of all features. \\ \bottomrule
\end{tabular}%
}
\caption{Examples from our dataset: shortened user persona descriptions, context factors, and associated contextual preferences.}
\label{tab:dataset-examples}
\end{table}
\renewcommand{\arraystretch}{1.0}

\section{Types for Context Factors, Preferences, and Tasks}
\label{appendix:types}

We define taxonomies context factor, preference, and task types. Summarized in \autoref{tab:taxonomy_summary}.

\paragraph{Context Factors}
For context factors, we manually annotated a sample of 256 random requests from WildBench~\citep{lin2024wildbench} as these represent realistic requests that human users ask to LLMs.
This resulted in a taxonomy of 8 context factor types.

\paragraph{Preference Types}
For preference types, we adapted the multiple preference facets described by~\citet{lee2024aligning}. Specifically, we adapted the sub-dimensions under the "style", "informativeness", and "harmlessness" dimensions. 
We did not include the "background knowledge" dimension as the sub-dimensions were less explicit about how the user's preferences would change. 
This resulted in 16 preference types.

\paragraph{Task Types}
To guide the LLM to consider diverse tasks that user may perform with the help of other LLMs, we also created a taxonomy of task types.
We referenced prior work that analyzed the most common tasks where real-world users are employing LLMs~\citep{tamkin2024clio, zao2024people}.
We combined the list of tasks described in these work by removing duplicates and combining similar or related tasks.
This resulted in 12 task types.

\begin{table}[t]
\centering
\resizebox{\textwidth}{!}{
\begin{tabular}{ll}
\toprule
\textbf{Dimension} & \textbf{Types}\\
\midrule
Context Factors & Person/Group, Organization/Institution, Object/Artifact, Content/Media, \\
                & Tool/Technology, Location/Place, Process/Activity, Event/Time \\[0.5em]

Preference Types & \textit{Style}: Formality, Clarity, Conciseness, Vividness, Format, Tone \\
               & \textit{Informativeness}: Relevance, Depth, Creativity, Efficiency, Practicality \\
               & \textit{Harmlessness}: Accuracy, Morality, Safety, Sensitivity, Trustworthiness \\[0.5em]

Task Types & Information seeking, Learning, Reasoning, Planning, Content creating, \\
           & Communicating, Writing \& editing, Coding \& debugging, Problem solving, \\
           & Data analysing, Advise seeking, Brainstorming \\
\bottomrule
\end{tabular}
}
\caption{Summary of Taxonomies for Context Factors, Contextual Preference, and Tasks.}
\label{tab:taxonomy_summary}
\end{table}

\section{Data Validation}
\label{appendix:validation}

\begin{figure}[h]
    \centering
    \includegraphics[width=\textwidth,height=0.98\textheight,keepaspectratio]{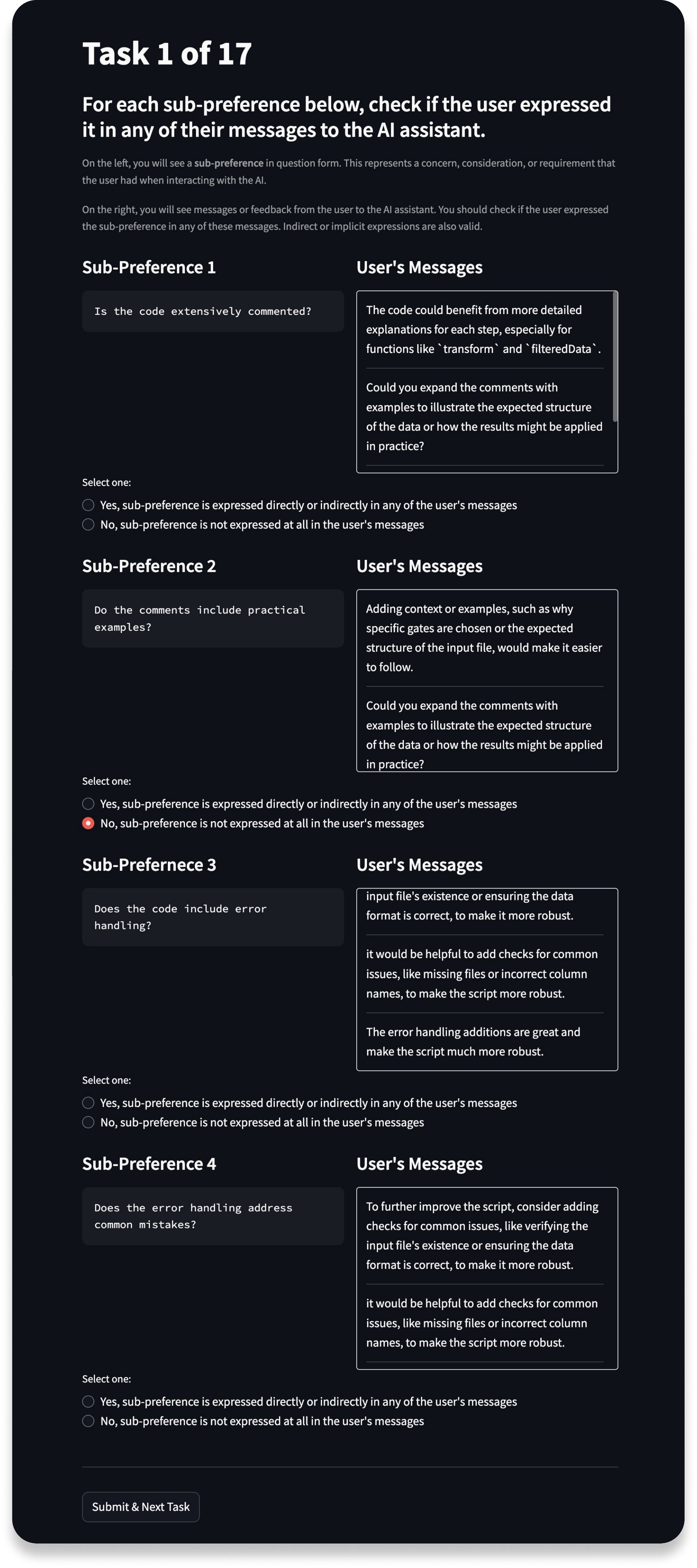}
    \caption{Annotation interface for data validation where human annotators were asked to verify whether a preference was expressed in the user's messages or not.}
    \label{fig:validation-interface}
\end{figure}

\paragraph{Human Validation}
Our dataset consists of 252 personas, where a series of interaction sessions was synthesized for each persona.
We performed human validation on each of these 252 interaction session series.
For each series, we created four validation tasks: three tasks to validate that each instance-relevant preference (i.e., $p_\text{current}$, $p_\text{contrast}$, $p_\text{prior}$) is expressed in the prior interaction sessions, and one task to validate that $p_\text{current}$ is not expressed in the current request, $u_\text{current}$.
This resulted in a total of 1008 tasks, where each task was performed by two annotators.
We recruited a total of 252 human annotators, where each annotator passed a qualification task and performed 16 tasks. 
On average, annotators spent approximately 30 minutes completing all the tasks and were compensated with £4.50. 
Using our annotation interface (\autoref{fig:validation-interface}), the annotators were presented with each entry from a preference's checklist, and either a list of user messages or only the user's request. 
Annotators were asked to mark whether the messages or request expressed or hinted at the checklist entry or not.
We asked for annotations on each checklist entry---rather than on the overall preference---to validate that each aspect of a preference was expressed (or not) in the relevant messages.
We then manually revised any messages were at least one annotator marked that any checklist entry was not expressed in the messages (or was expressed in the user's current request).
Additionally, for the task where annotators validated the current request $u_\text{current}$, we also asked annotators to rate how realistic that request was. 
Specifically, annotators were asked to consider whether it is realistic for a user to ask this request to an AI assistant (e.g., ChatGPT, Gemini, Perplexity, Claude, DeepSeek, etc.) by providing a rating between  1 (not realistic at all) to 5 (extremely realistic).

\paragraph{Message Extraction}
To reduce the cognitive load on the annotators and the task time, we used an LLM to first extract the message fragments relevant to a preference from the prior interaction sessions.
Specifically, for each preference, we first prompted \texttt{gpt-4o-2024-11-20} with the preference's checklist and the relevant dialogues, and asked it to extract the messages that could hint or express each entry in the preference.
Prompt is in \autoref{fig:message-extract-prompt}.

\begin{figure}[t]
  \centering
  \begin{minipage}{0.328\textwidth}
    \centering
    \includegraphics[width=\textwidth]{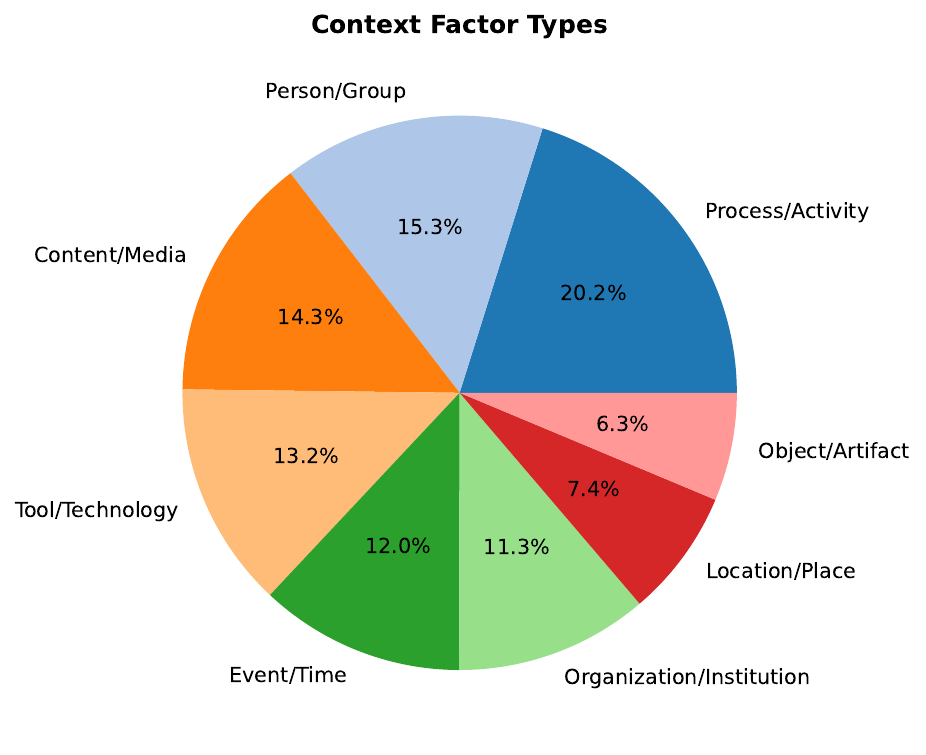}
  \end{minipage}
  \hfill
  \begin{minipage}{0.328\textwidth}
    \centering
    \includegraphics[width=\textwidth]{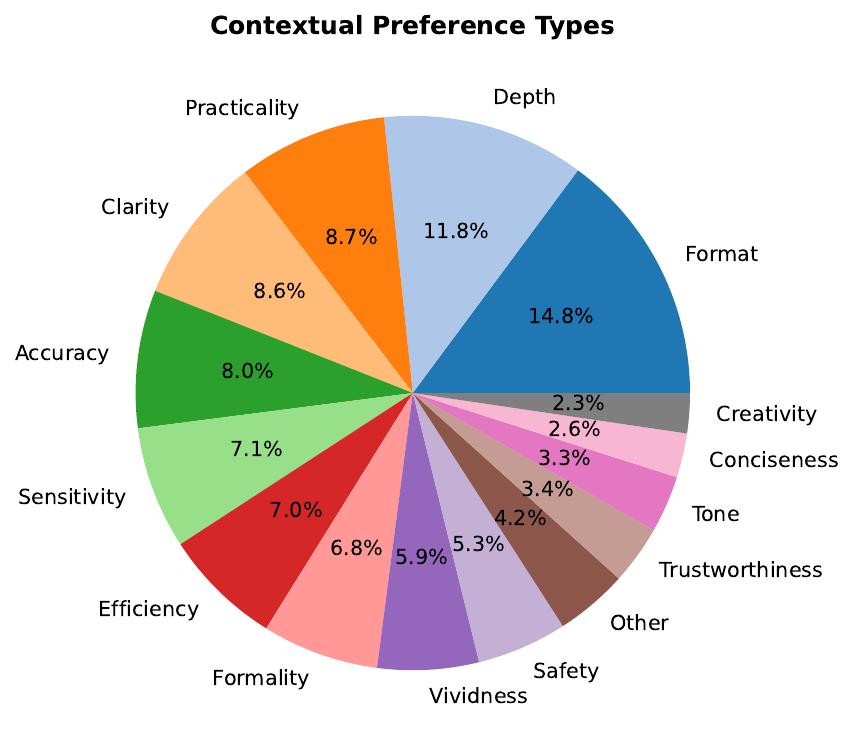}
  \end{minipage}
  \hfill
  \begin{minipage}{0.328\textwidth}
    \centering
    \includegraphics[width=\textwidth]{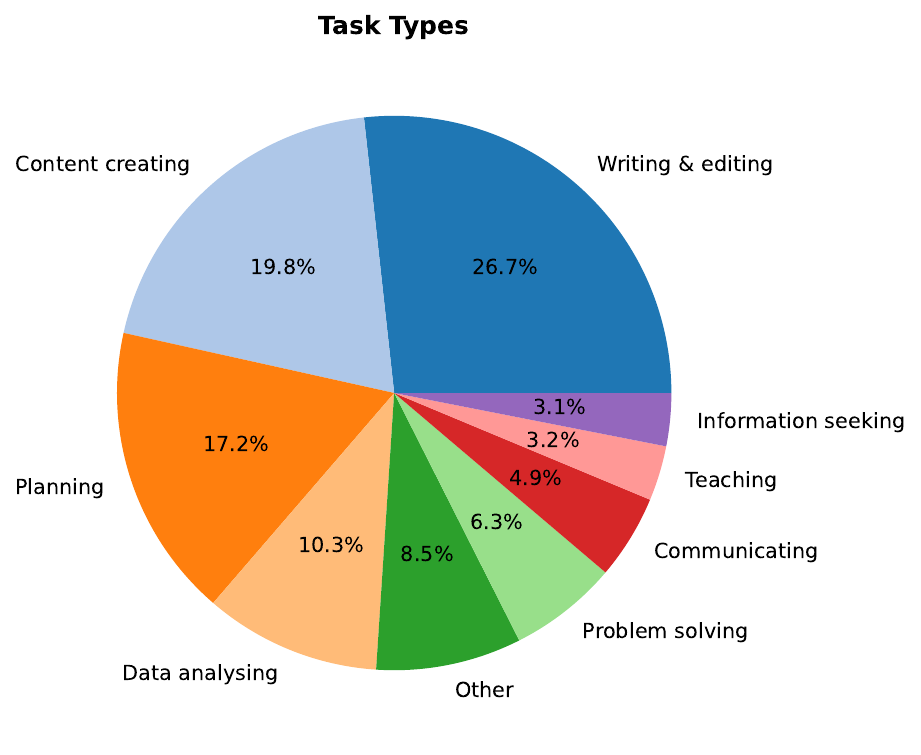}
  \end{minipage}
  \vspace{-10pt}
  \caption{Distribution of the types of context factors, contextual preferences, and tasks included in our dataset.}
  \label{fig:dataset-distribution}
\end{figure}

\section{Additional Experiment Details and Results}
\label{appendix:additional_results}

Full details for models evaluated in our experiments are shown in \autoref{tab:model-details}.

\renewcommand{\arraystretch}{1.25}
\begin{table}[b]
\centering
\resizebox{0.8\textwidth}{!}{
\begin{tabular}{@{}llll@{}}
\toprule
Model             & Version        & API            & Additional Parameters   \\ \midrule
GPT-4o            & 2024-08-07     & OpenAI         & temperature=0                \\
o3-mini           & 2025-01-31     & OpenAI         & (* temperature is not supported.) \\
Claude 3.7 Sonnet &
  20250219 &
  Amazon Bedrock &
  \begin{tabular}[c]{@{}l@{}}temperature=1 (* only supported setting.)\\thinking=\{type: "enabled", budget\_token: 1024\}\end{tabular} \\
Claude 3.5 Sonnet & 20240620       & Amazon Bedrock & temperature=0                \\
Llama 3.1 405B    & Instruct Turbo & Together AI    & temperature=0                \\
Mistral 7B        & Instruct v0.3  & Together AI    & temperature=0                \\
Qwen2.5 72B       & Instruct Turbo & Together AI    & temperature=0                \\
DeepSeek-R1       &                & Together AI    & temperature=0                \\
Gemini 2.0 Flash Thinking & exp-01-21 & Vertex AI & temperature=0 \\
Gemini 2.0 Pro    & exp-02-05      & Vertex AI      & temperature=0                \\ \bottomrule
\end{tabular}
}
\caption{Full details for models tested in our experiments.}
\label{tab:model-details}
\end{table}
\renewcommand{\arraystretch}{1.0}

\begin{figure}[b!] 
    \begin{center}
    \includegraphics[width=0.85\textwidth]{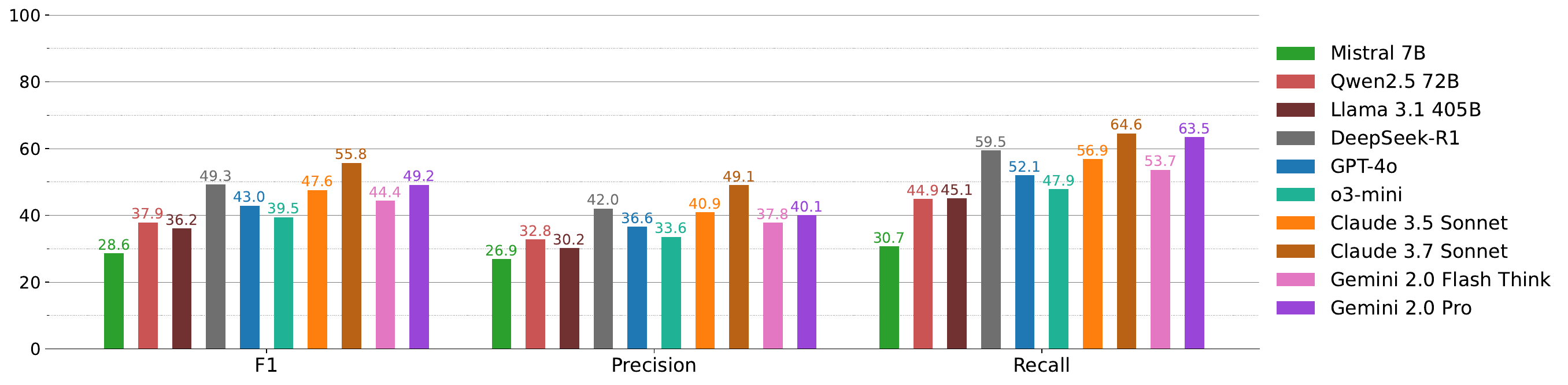}
    \includegraphics[width=0.85\textwidth]{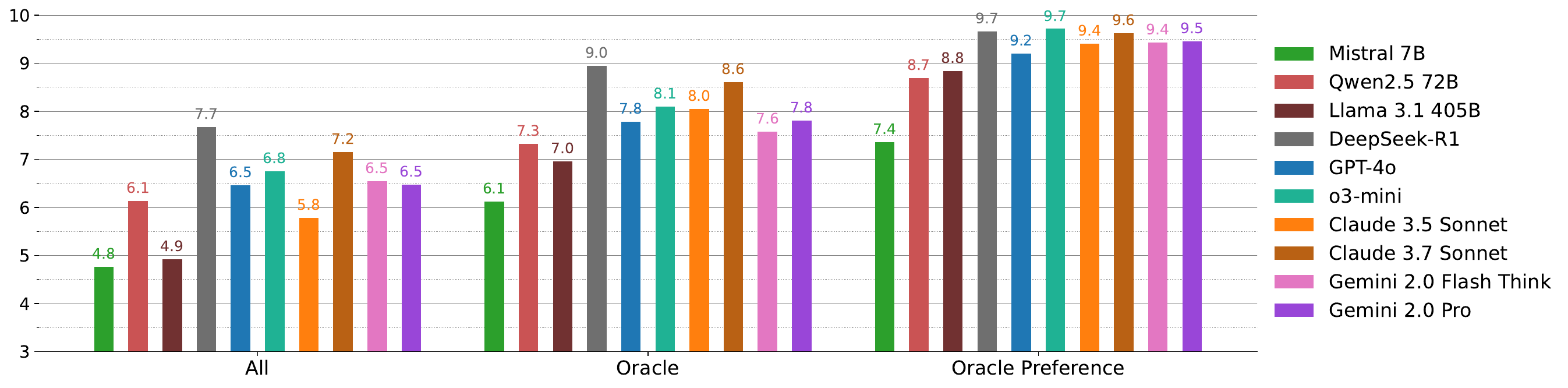}
    \caption{(Top) Average F1, precision, and recall for all models in the inference task in all instances. (Bottom) Average alignment score for all models in the generation task averaged across all instances, for the oracle setting, and for the oracle preference setting.}
    \label{fig:results-visualization}
    \end{center} 
\end{figure}

\subsection{Visualization of Results}

\autoref{fig:results-visualization} presents visualizations of the results for all the evaluated models in the Inference and Generation tasks.

\subsection{Summary Results}

\autoref{tab:inference-summary} and \autoref{tab:generation-summary} presents the full results for the inference task with the summaries.
Most open-source models appear to present a significant increase in performance with the summaries.
Interestingly, the performance of certain models decreases slightly with the summaries, which appear to mostly be the models with reasoning capabilities.
Additionally, we see a similar trend in performance regarding the instance types, where, in general, performance is highest in the Changing instances and lowest in the Contrastive instances---reflecting that models may also be focusing on the most recent summaries as they did with the interaction sessions.
Interestingly, we observe that the summaries help each model in different task.
Specifically, we observe that Mistral 7B, Qwen2.5 72B, and Llama 3.1 405B see significant increases in inference performance with the summaries, but not in the generation task.
While Claude 3.5 Sonnet and Gemini 2.0 Pro see minimal gains in the inference task with summaries, but their performance in the generation task increases significantly.

\begin{table}[b!]
\centering
\resizebox{0.9\textwidth}{!}{%
\begin{tabular}{@{}lccc|ccccccccc@{}}
\toprule
\textbf{} &
  \multicolumn{3}{c|}{\textbf{All Instances}} &
  \multicolumn{3}{c}{\textbf{Consistent}} &
  \multicolumn{3}{c}{\textbf{Contrastive}} &
  \multicolumn{3}{c}{\textbf{Changing}} \\
\textbf{Model} &
  \textbf{P} &
  \textbf{R} &
  \textbf{F1} &
  \textbf{P} &
  \textbf{R} &
  \textbf{F1} &
  \textbf{P} &
  \textbf{R} &
  \textbf{F1} &
  \textbf{P} &
  \textbf{R} &
  \textbf{F1} \\ \midrule
Mistral 7B        & 36.0 & 49.8 & 41.8 & 35.1 & 49.1 & 40.9 & 33.4 & 44.2 & 38.0 & 39.8 & 56.3          & 46.6 \\
Qwen2.5 72B       & 37.2 & 55.3 & 44.5 & 37.7 & 56.2 & 45.1 & 37.3 & 52.8 & 43.7 & 36.7 & 56.8          & 44.6 \\
Llama 3.1 405B    & 35.4 & 54.7 & 43.0 & 36.4 & 56.9 & 44.4 & 34.4 & 51.5 & 41.2 & 35.3 & 55.7          & 43.2 \\
DeepSeek-R1       & 40.1 & 60.0 & 48.0 & 41.1 & 61.9 & 49.4 & \underline{41.3} & \underline{58.1} & \underline{48.3} & 37.8 & 59.9          & 46.3 \\
GPT-4o            & 37.7 & 58.5 & 45.9 & 37.1 & 56.2 & 44.7 & 36.6 & 56.2 & 44.3 & 39.6 & 63.1          & 48.6 \\
o3-mini           & 34.7 & 50.7 & 41.2 & 35.4 & 52.0 & 42.1 & 33.2 & 49.4 & 39.8 & 35.4 & 50.8          & 41.7 \\
Claude 3.5 Sonnet & \underline{42.1} & 61.9 & \underline{50.1} & \underline{43.0} & \underline{63.9} & \underline{51.4} & 40.5 & 56.8 & 47.3 & \underline{42.8} & 65.0          & 51.6 \\
Claude 3.7 Sonnet &
  \textbf{45.4} &
  \textbf{64.4} &
  \textbf{53.3} &
  \textbf{45.6} &
  \textbf{64.0} &
  \textbf{53.3} &
  \textbf{45.1} &
  \textbf{63.0} &
  \textbf{52.6} &
  \textbf{45.6} &
  \underline{66.1} &
  \textbf{54.0} \\
Gemini 2.0 Flash Thinking & 37.3 & 53.9 & 44.1 & 36.4 & 52.7 & 43.0 & 37.1 & 52.4 & 43.4 & 38.6 & 56.7 & 45.9 \\
Gemini 2.0 Pro    & 37.1 & \underline{62.1} & 46.4 & 35.1 & 62.5 & 45.0 & 33.8 & 55.0 & 41.9 & 42.5 & \textbf{68.9} & \underline{52.5} \\ \bottomrule
\end{tabular}
}
\caption{Precision (P), Recall (R), and F1 score for all models on the inference task with summaries. Best results in each column are \textbf{bold}-faced and second best results are \underline{underlined}.}\label{tab:inference-summary}
\end{table}
\begin{table}[b!]
\centering
\resizebox{0.7\textwidth}{!}{%
\begin{tabular}{@{}lcccc@{}}
\toprule
\textbf{}                  & \multicolumn{4}{c}{\textbf{Judge Scores}}      \\
\textbf{Model}          & \multicolumn{1}{c|}{\textbf{All}}  & \textbf{Consistent} & \textbf{Contrastive} & \textbf{Changing} \\ \midrule
Mistral 7B        & \multicolumn{1}{c|}{5.02} & 4.78 & 4.50 & 5.79 \\
Qwen2.5 72B       & \multicolumn{1}{c|}{6.41} & 6.23 & 5.75 & 7.25 \\
Llama 3.1 405B    & \multicolumn{1}{c|}{5.05} & 4.77 & 4.50 & 5.90 \\
DeepSeek-R1    & \multicolumn{1}{c|}{\textbf{7.70}} & \textbf{7.75}       & \textbf{7.33}        & \underline{8.01}              \\
GPT-4o            & \multicolumn{1}{c|}{6.65} & 6.29 & 6.13 & 7.54 \\
o3-mini           & \multicolumn{1}{c|}{6.92} & 6.80 & 6.49 & 7.46 \\
Claude 3.5 Sonnet & \multicolumn{1}{c|}{6.81} & 6.64 & 6.36 & 7.44 \\
Claude 3.7 Sonnet & \multicolumn{1}{c|}{\underline{7.57}} & \underline{7.46} & \underline{7.32} & 7.94 \\
Gemini 2.0 Flash Thinking & \multicolumn{1}{c|}{6.35} & 5.94 & 5.74 & 7.37 \\
Gemini 2.0 Pro & \multicolumn{1}{c|}{7.35}          & 7.03                & 6.80                 & \textbf{8.24}     \\ \bottomrule
\end{tabular}
}
\caption{Preference alignment scores on the generation task with summaries.}
\label{tab:generation-summary}
\end{table}

\subsection{Inference with \matcherName{}}

\autoref{fig:results-pref-matcher} shows the full results in the inference task when assessed with \matcherName{}.
As shown, inference performance when assessed with \matcherName{} and with GPT-4o generally only show marginal differences, without one of the models skewing towards providing higher or lower scores.

\begin{table}[]
\resizebox{\textwidth}{!}{%
\begin{tabular}{@{}lccc|ccccccccc|ccc@{}}
\toprule
\textbf{} &
  \multicolumn{3}{c|}{\textbf{All Instances}} &
  \multicolumn{3}{c}{\textbf{Consistent}} &
  \multicolumn{3}{c}{\textbf{Contrastive}} &
  \multicolumn{3}{c|}{\textbf{Changing}} &
  \multicolumn{3}{c}{\textbf{Oracle Setting}} \\
\textbf{Model} &
  \textbf{P} &
  \textbf{R} &
  \textbf{F1} &
  \textbf{P} &
  \textbf{R} &
  \textbf{F1} &
  \textbf{P} &
  \textbf{R} &
  \textbf{F1} &
  \textbf{P} &
  \textbf{R} &
  \textbf{F1} &
  \textbf{P} &
  \textbf{R} &
  \textbf{F1} \\ \midrule
Mistral 7B & 27.0 & 29.7 & 28.3 & 22.4 & 26.1 & 24.1 & 23.0 & 27.6 & 25.1 & 35.8 & 35.7 & 35.7 & 46.0 & 58.6 & 51.5 \\
Qwen2.5 72B & 33.0 & 46.5 & 38.6 & 29.8 & 44.2 & 35.6 & 27.2 & 38.8 & 32.0 & 42.0 & 56.4 & 48.1 & 58.8 & 75.5 & 66.1 \\
Llama 3.1 405B & 28.7 & 44.2 & 34.8 & 24.9 & 38.3 & 30.2 & 25.9 & 40.5 & 31.6 & 35.4 & 53.8 & 42.7 & 56.9 & 75.8 & 65.0 \\
DeepSeek-R1 & \underline{41.2} & 59.6 & \underline{48.7} & \underline{42.9} & 62.5 & \underline{50.9} & \underline{40.9} & \underline{58.9} & \underline{48.3} & 39.6 & 57.3 & 46.8 & 60.7 & 80.7 & 69.3 \\
GPT-4o & 34.9 & 52.9 & 42.0 & 34.0 & 49.4 & 40.3 & 30.7 & 47.3 & 37.2 & 39.9 & 62.1 & 48.6 & 56.1 & 76.5 & 64.8 \\
o3-mini & 32.2 & 50.0 & 39.2 & 31.8 & 49.3 & 38.7 & 29.1 & 47.5 & 36.1 & 35.7 & 53.2 & 42.8 & 47.9 & 69.0 & 56.6 \\
Claude 3.5 Sonnet & 40.6 & 58.3 & 47.9 & 40.2 & 57.2 & 47.2 & 37.4 & 54.1 & 44.2 & \underline{44.1} & 63.8 & \underline{52.2} & \underline{61.6} & 80.8 & \underline{69.9} \\
Claude 3.7 Sonnet & \textbf{48.5} & \textbf{65.6} & \textbf{55.7} & \textbf{50.9} & \textbf{67.5} & \textbf{58.0} & \textbf{47.7} & \textbf{62.5} & \textbf{54.1} & \textbf{46.9} & \underline{66.7} & \textbf{55.1} & \textbf{69.3} & \textbf{84.4} & \textbf{76.1} \\
Gemini 2.0 Flash Think & 36.5 & 53.7 & 43.4 & 35.7 & 54.0 & 43.0 & 33.7 & 49.9 & 40.3 & 39.9 & 57.2 & 47.0 & 57.2 & 73.9 & 64.5 \\
Gemini 2.0 Pro & 39.1 & \underline{63.0} & 48.3 & 37.6 & \underline{63.0} & 47.1 & 37.7 & 57.9 & 45.6 & 42.0 & \textbf{68.1} & 52.0 & 61.0 & \underline{81.5} & 69.8 \\ \bottomrule
\end{tabular}
}
\caption{Precision (P), Recall (R), and F1 score for all models in the inference task when assessed with \matcherName{}.}
\label{tab:inference-matcher}
\end{table}

\begin{figure}[t]
  \centering
  \begin{minipage}{0.497\textwidth}
    \centering
    \includegraphics[width=\textwidth]{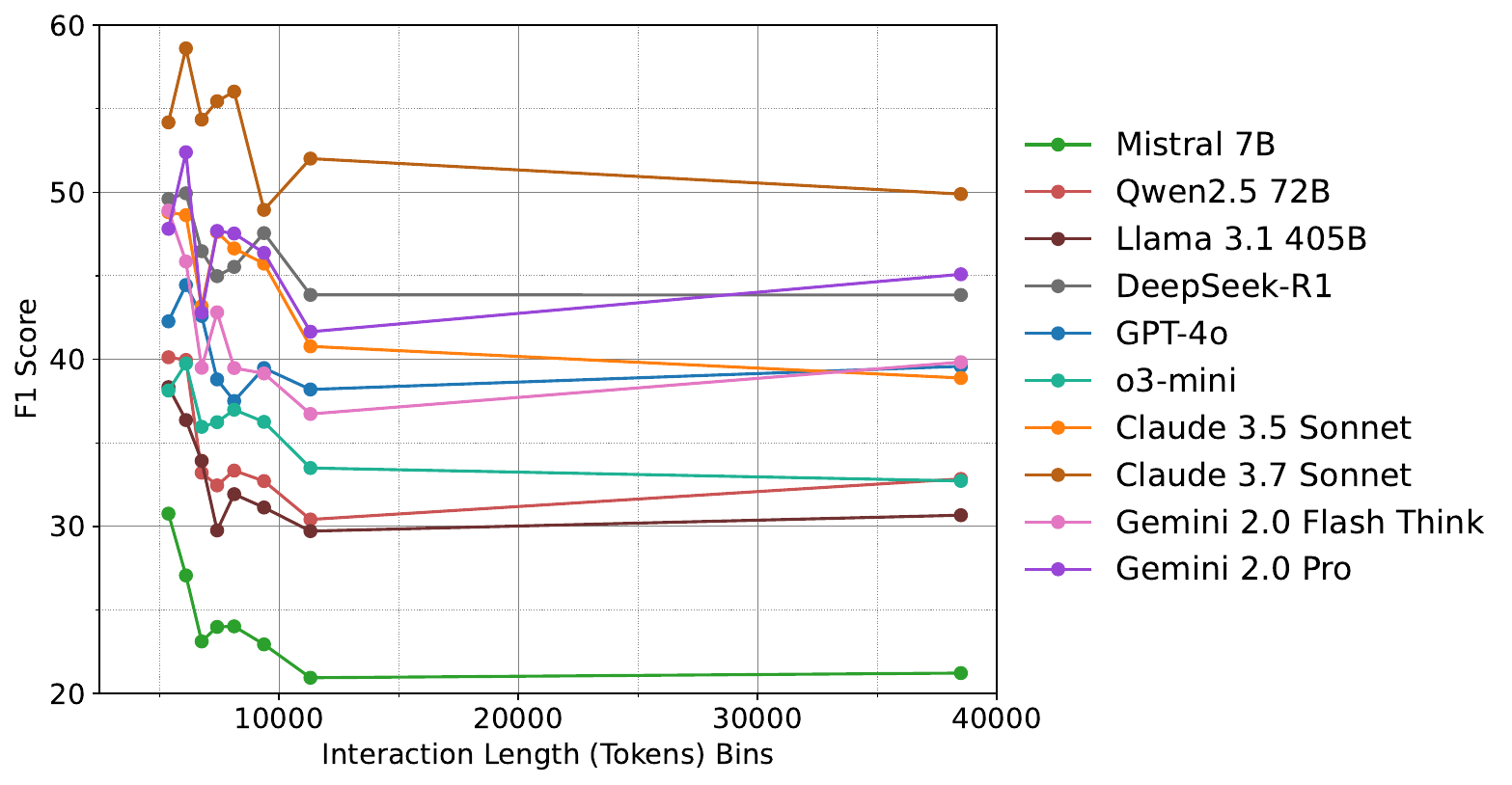}
  \end{minipage}
  \hfill
  \begin{minipage}{0.497\textwidth}
    \centering
    \includegraphics[width=\textwidth]{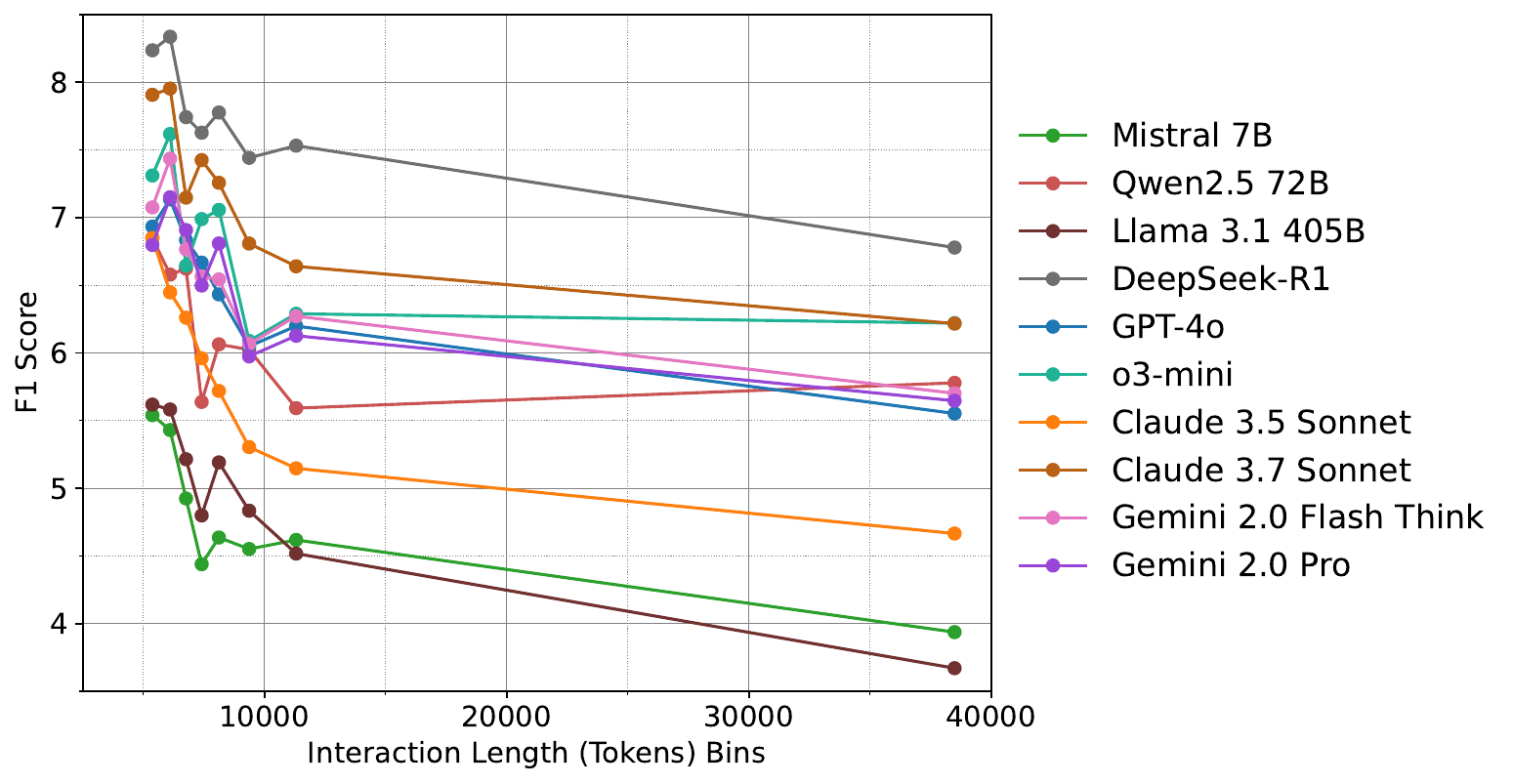}
  \end{minipage}
  \caption{Average result for each model in the inference task (left) and generation task (right) according to the maximum token length of the instances.}
  \label{fig:interaction-length}
\end{figure}

\subsection{Performance against Length of Prior Interaction Sessions}

\autoref{fig:interaction-length} shows the average results of each model in the inference and generation task depending on the token length of the previous interaction sessions in the instances.
Specifically, we divided all the instance data into equally-sized groups (bins). Each bin contains instances with lengths larger than the maximum length in the previous bin, but less than the minimum length in the next bin. Then, we averaged model performance for all data in each bin. This allows us to observe that each bin represents model performance at a distinct range of lengths.
As seen from the results, we observe a general trend where model performance in both tasks decreases with the length of the prior interaction sessions---which coincides with findings in similar work \citep{wu2024longmemeval, zhao2025llms}.
This suggests that, with even longer interaction sessions, models will struggle to infer users' personalized and contextual preferences from interactions.

\begin{figure}[b!]
    \centering
    \includegraphics[width=0.9\textwidth]{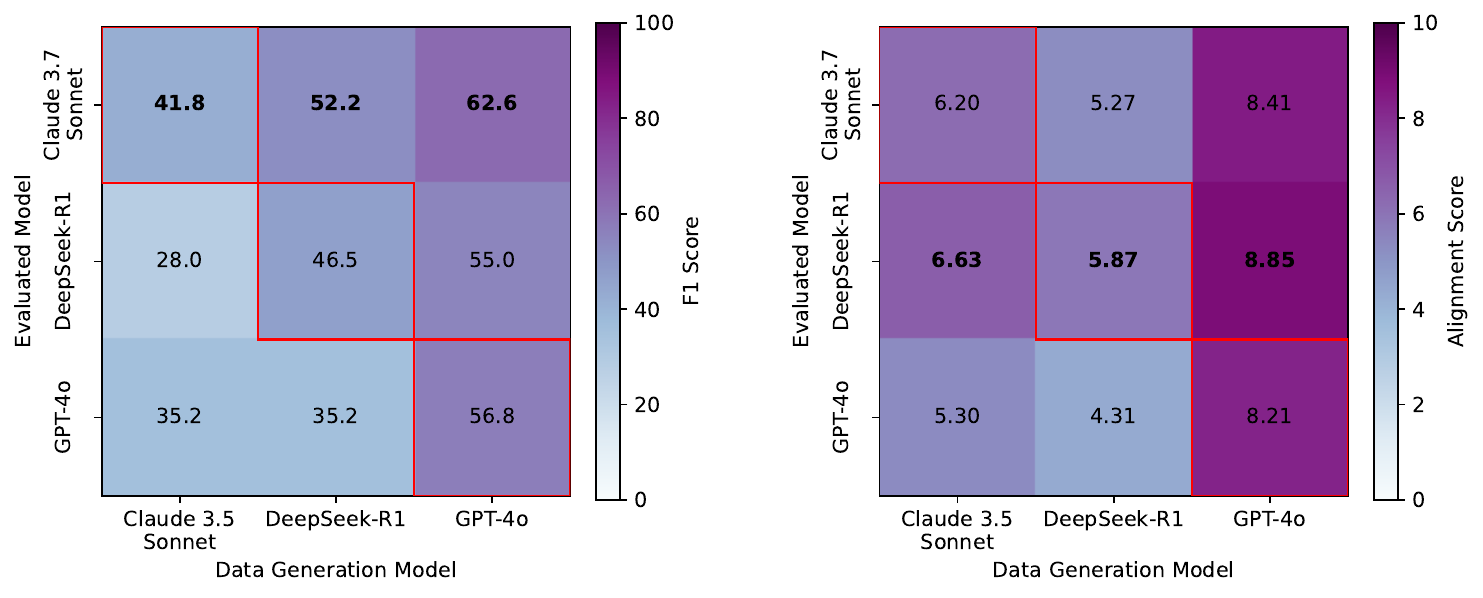}
    \caption{Inference and generation performance of tested models on datasets synthesized by other tested models. Red‑highlighted squares are evaluations on data generated by the same model (or a model of the same family).}
    \label{fig:self_enhancement}
\end{figure}

\subsection{Analyzing Possible Self-Enhancement Bias}
\label{appendix:self-enhancement}

As \benchmark{} context factor-preferences were generated by Claude 3.5 Sonnet, Claude 3.7 Sonnet's superior inference performance raises the possibility that implicit biases trained into these models were reflected in the data.
To test this, we synthesized small datasets using three high-performing models---Claude 3.7 Sonnet, GPT-4o, and DeepSeek-R1---based on the 64 persona profiles in our benchmark that produced the lowest average performance across all models .
For each persona and model, we used our synthesis pipeline to generate Consistent, Contrastive, and Changing instances---leading to small datasets of 192 instances for each model.
Then, we evaluated each model on its own and the other models' datasets---for Claude 3.7 Sonnet, we instead used the original \benchmark{} instances for the 64 personas.

\autoref{fig:self_enhancement} shows minimal evidence of higher performance on self-generated data, indicating that Claude 3.7 Sonnet's strong results in our benchmark reflect genuine capability rather than dataset bias.
Performance patterns in all the datasets were consistent with those in our benchmark: in inference, Claude 3.7 Sonnet performed best, then DeepSeek‑R1, and then GPT‑4o; in generation, DeepSeek‑R1 led, with Claude 3.7 Sonnet second, and GPT‑4o last.

\begin{figure*}[h]
\begin{tcolorbox}[width=\linewidth, fontupper=\scriptsize]

\textbf{User Persona Profile} \\

\hrule
\bigskip

\textbf{System Prompt} 

\begin{Verbatim}[breaklines, fontsize=\tiny]
Your task is to design user personas.

For each persona, you'll receive a basic persona description and the following attributes: 
- Occupation
- Career Level
- Personality Traits
{personality_traits_list}
- Personal Values
{personal_values_list}
- Decision-Making Style
{decision_making_style_list}

Using these details, write a six-sentence expanded persona description following the content of each sentence:
1. Given Name (think of culturally diverse names), Age, Gender/Pronouns, Occupation, Career Level
2. Gain: Wants, needs, and measures of success.
3. Pain: Fears, frustrations, and obstacles.
4. Think and Feel: What really counts, their major preoccupations, worries, and aspirations.
5. Hear and See: How the persona views their environment, their friends, co-workers, etc. What others say about the persona.
6. Say and Do: How does the persona behave towards others, what do they do in their daily and work life.

Follow these guidelines:
- Ensure that your descriptions are balanced by exploring both the positives/strengths and negatives/weaknesses of the persona.
- Substitute the phrases which directly mention the personas' traits, such as due to her introverted nature", "his high neuroticism", or "her conceptual style", with other phrases that reveal the traits implicitly.
- You should show not tell, but keeping the tone clear and direct. Avoid emotional or subjective language and exaggeration.
- If any traits appear to conflict, address how these can coexist within the same individual and how they manifest in different contexts, reflecting the complexity of human behavior. For example, a person with a "Low Conscientiousness" personality may adopt a more "Directive" style at work to avoid making mistakes.
- Make sure that all the sentences are logically connected to each other and the basic persona description. Make the description natural and coherent, narrating the persona's behavior and backstory.
- Ensure that each sentence reveals new details and information, instead of repeating content similar to the other sentences.
- Think of broader, more abstract tasks that this persona might undertake throughout their life based on the basic persona description or seed. Consider mentioning various tasks that could span different domains that the persona might engage in, such as work, social life, or personal hobbies.

Please return your response in the following JSON format:

```json
[
    {{
      "description": "<6-sentence persona description illustrating their thoughts, behavior, actions, and backstory>",
      "occupation": "<persona's occupation>"
    }},
    {{
      "description": "<6-sentence persona description illustrating their thoughts, behavior, actions, and backstory>",
      "occupation": "<persona's occupation>"
    }},
    ...
]
```
\end{Verbatim}

\hrule
\bigskip

\textbf{User Prompt}

\begin{Verbatim}[breaklines, fontsize=\tiny]
## List of Basic User Descriptions

{seed_personas}
\end{Verbatim}

\end{tcolorbox}
\caption{Prompt to generate a user persona's profile from a seed description and a set of attributes.}
\label{fig:profile-generate-prompt}
\end{figure*}

\begin{figure*}[ht]
\begin{tcolorbox}[width=\linewidth, fontupper=\scriptsize]

\textbf{Measure Preference Match} \\

\hrule
\bigskip

\textbf{System Prompt}

\begin{Verbatim}[breaklines, fontsize=\tiny]
In this task, you will be presented with a **evaluation checklist** and a **preference**. The preference describes an aspect of AI outputs that should be evaluated. The checklist contain questions that are used to evaluate more specific or fine-grained aspects of the AI outputs.

Your task is to determine whether each entry in the checklist is **covered** by the preference. **Covered** means that the preference and the checklist entry will evaluate the same or similar aspects of an AI output, even if they use different wording or phrasing. A preference and checklist entry can refer to the subject matter differently (e.g., "response" vs "report") but still evaluate the same aspects. Ignore differences in wording and focus on the underlying aspects being evaluated.

For each checklist entry, you can choose one of the following options:

1. **Fully Covered**: The preference fully covers or evaluates the checklist entry. Evaluating on the preference will also evaluate the checklist entry.

2. **Partially Covered**: The preference partially covers or evaluates the checklist entry. Evaluating on the preference may evaluate some aspects of the checklist entry.

3. **Not Covered**: The preference does not cover or evaluate the checklist entry.

### Output Format

Provide your results in the following JSON format. Ensure to include the code block markers (```).

```json
{{
    "results": [
      {{
        "index": <index of the entry in the checklist in the first pair>,
        "entry": <an entry from the checklist in the first pair>,
        "label": <"Fully Covered" / "Partially Covered" / "Not Covered">
      }},
      {{
        "index": <index of the entry in the checklist in the first pair>,
        "entry": <an entry from the checklist in the first pair>,
        "label": <"Fully Covered" / "Partially Covered" / "Not Covered">
      }},
      ...
    ]
}}
```

---

### Examples

{examples}
\end{Verbatim}

\hrule
\bigskip

\textbf{User Prompt}

\begin{Verbatim}[breaklines, fontsize=\tiny]
#### Preference

{preference}

#### Checklist

{checklist}
\end{Verbatim}

\end{tcolorbox}
\caption{Prompt to measure how much a checklist matches with a preference.}
\label{fig:preference-match-prompt}
\end{figure*}
\begin{figure*}[h]
\begin{tcolorbox}[width=\linewidth, fontupper=\scriptsize]

\textbf{Preference Decompose to Checklist} \\

\hrule
\bigskip

\textbf{System Prompt}

\begin{Verbatim}[breaklines, fontsize=\tiny]
## Your Objective

Your task is to help judge how well an AI Assistant's response satisfies a given preference by creating an evaluation checklist from the preference. Here, a **preference** refers to a requirement, guideline, or principle that a user considers when assessing the quality of an AI Assistant's response.

## Task Details

Your task is to come up with an evaluation checklist list for a given preference. This evaluation checklist should be a list of questions that ask whether or not specific aspects contained within a preference were met by an AI assistant’s response.

Aspects covered by your checklist should be explicitly stated in the preference. You should try to be concise and avoid including unnecessary entries in your checklist that were not contained in the preference.

Checklist questions should:
- **Be answerable by ’yes’ or ’no’**, with ’yes’ meaning that the response successfully met the corresponding requirement.
- **Be comprehensive, but concise**, meaning that all aspects that are directly relevant to the preference should be represented by a question, but only questions that are very clearly relevant should be included.
- **Be precise**, meaning that checklist questions should avoid vague wording and evaluate specific aspects of a response, directly using the phrasing of the preference where appropriate. Avoid checklist entries that introduce new content that is not included in the preference.

You should always analyse the preference before providing the evaluation checklist. The checklist should contain **at most 4 entries**.

## Response Format

**Analysis**

<Explain your analysis of the preference here.>

**Checklist**

```json
{{
    "checklist": [
        <Each entry of the checklist in a separate line>,
        <Each entry of the checklist in a separate line>
      ...
    ]
}}
```

---

## Examples

{examples}
\end{Verbatim}

\hrule
\bigskip

\textbf{User Prompt}

\begin{Verbatim}[breaklines, fontsize=\tiny]
**Preference**

{preference}
\end{Verbatim}

\end{tcolorbox}
\caption{Prompt to decompose contextual preferences into checklists.}
\label{fig:preference-decompose-prompt}
\end{figure*}
\begin{figure*}[h]
\begin{tcolorbox}[width=\linewidth, fontupper=\scriptsize]

\textbf{Judge Response Quality} \\

\hrule
\bigskip

\textbf{System Prompt}

\begin{Verbatim}[breaklines, fontsize=\tiny]
## **Your Objective**

You are a critical and meticulous evaluator. You will be presented with a user's request to an AI assistant and the AI's response to the user. Your task is to evaluate whether the AI assistant's response satisfied the user's **personal preference**. To help you evaluate the responses on the preference, you will also be provided with an **evaluation checklist** that decomposes the preference into specific questions.

### **Evaluation Preference and Checklist**

**Preference**: "{preference}"

**Evaluation Checklist**:
{checklist}

## **Instructions**

You should write down your analysis and assessment on how well the AI assistant's response satisfies each item in your checklist. You should follow these considerations:

- Walk through each checklist item and summarize the response's "strengths" and "weaknesses" regarding that checklist item. 
- For each checklist item, you should consider whether the checklist item was satisfied or dissatisfied.
- Avoid considering aspects that are not included in the checklist. Focus only on the evaluation checklist. Ensure that your persona profile does not influence your evaluation.
- You should then return a score in the range of 1~10, where 1 means a response is very poor and 10 means the response is completely perfect.

Ensure that you follow the format given below. Avoid adding additional content that is not included in the format below.

---

## **Output Format**

### Evaluation of AI Assistant's Response

1. **<Checklist item 1>**: <Detailed analysis and evaluation of the AI assistant's response on the first item in the checklist>

2. **<Checklist item 2>**: <Detailed analysis and evaluation of the AI assistant's response on the second item in the checklist>

...

### Evaluation Score

<Return your numeric score in the range of 1~10>
\end{Verbatim}

\hrule
\bigskip

\textbf{User Prompt}

\begin{Verbatim}[breaklines, fontsize=\tiny]
### User's Request

{user_request}

### AI Assistant's Response

{ai_response}
\end{Verbatim}

\end{tcolorbox}
\caption{Prompt to evaluate the quality of an LLM's response on a given preference and checklist.}
\label{fig:response-judge-prompt}
\end{figure*}
\begin{figure*}
\begin{tcolorbox}[width=\linewidth, fontupper=\scriptsize]
\textbf{Context Factors Generator} \\

\hrule
\bigskip

\textbf{System Prompt}

\begin{Verbatim}[breaklines, fontsize=\fontsize{5}{6}\selectfont]
### **Your Objective:**

You will be provided with a **user persona**. This user will interact with a conversational AI assistant (e.g., ChatGPT) in diverse **scenarios**. In each scenario, the user will ask the AI assistant to help them with a **task** in their worklife. Consider that diverse **factors** exist within the user's world and environment that influence their behavior and expectations in various contexts. This means that, if a context factor is involved in the interactions the user has with the AI assistant, the user will have certain expectations about the AI assistant based on their experiences and understanding of the context factor.

Your goal is to imagine the following by being creative and imagining comprehensive and rich details about the user's world:
1. Identify **context factors** that exist within the user's work life and environment.
2. Imagine the **contextual preferences** that the user will consider when interacting with the AI assistant when each factor is involved---based on the user's knowledge or experiences with the factor in their environment.
You should return your final output in valid YAML format.

### **Definitions**:

**Scenario**:
Situation within a user's life that involves one or more factors within the user's world or environment that will influence how the user behaves and what they intent/expect in that scenario.

**Task**:
These are tasks, problems, assignments, or pieces of work that the user undergoes in the scenarios in their worklife. In particular, we are focused on the tasks where the user may need the help of a conversational AI assistant. Tasks should belong to one of the following types:
{task_type_list}

**Context Factors**:
These are specific and concrete elements within the user's world and environment that can be identified, described, and interacted with. These factors should be stable and significant, meaning that each factor will be present in various situations or scenarios in the user's work life. Consider diverse types of factors that the user interacts with in their work life. The factors should be realistic factors that exist in the actual real-world or fictional factors that are specific to the user's life. Factors should be one of the following types:
{factor_type_list}

**Contextual Preferences**:
These are guidelines, requirements, constraints, or principles that the AI assistant should align with in order to satisfy the user in given contexts (i.e., situations where a context factor is involved). These should be uniquely personal to the user and based on the user's personal life experiences with that context factor. You should imagine diverse types of preferences. You should create preferences that belong to one of the following types, which are organized into categories:
{preference_type_list}

### **Output Requirements**:

For each context factor within the user's world:
1. **Select a Factor Type**: Select a type of factor that you will create.
2. **Imagine the Background**: Create a unique, rich, and personal backstory that illustrates a specific and concrete factor in the user's world or environment with the selected factor type. The narrative should describe how the user typically interacts with this factor and the tasks where this factor can be involved, even as a minor presence. This narrative should also present significant previous experiences that the user had with this factor and how this formed the user's expectations around the factor. Finally, you should describe how this influences the user's intentions in situations that involve the context factor.
3. **Name the Context Factor**: You should now provide the specific name of the factor. Avoid using generic placeholder names. Instead, imagine realistic and specific names for the factor.
4. **Determine Task Types**: Based on the context factor and the background about the context factor, you should select possible tasks types from the given list. You should select types of tasks where the user needs the help of the AI assistant and where the context factor will be involved, even if it is a minor presence. You must select task types from the list. You can only select types from the list; avoid creating new types.
5. **Select a Preference Type**: Select the type of preference that you will create. You should not select the category but instead select from the given types within each category.
6. **Define the Contextual Preference**: Define the preference that encompasses the user's intentions and expectations for the AI assistant in the task types where the context factor is involved. This should not be commonsense knowledge or commonly held prefrences, but instead the preference should tie to the user-specific experiences. Ensure that anyone can understand the contextual preference by itself, without other information like the task or context factor. The preference should be clear, interpretable, and usable for external human evaluators or AI evaluators.

You should create **8** unique context factors, each with a backstory, a description, and a contextual preference.

**Create Contrastive Factors**: 
- The **last two (2)** context factors should be a **contrastive pair**. 
- A **contrastive pair** is two factors that are similar and have the same factor type. However, their contextual preferences must be incompatible, conflicting, or even contradictory to each other.
- Ensure that you satisfy the following requirements:
  - **Distinct Factors**: The contrastive factors should be clearly distinct from each other. Avoid creating entity pairs that are the same factor but with different traits. For example, avoid pairs where the only difference is the time of day, the version of the factor, or a trait of the factor has changed.
  - **Unique Preferences**: The preferences of the contrastive factors should be unique to the user's personal experiences. Avoid creating generic, commonsense, or universal preferences. Ensure that the preferences cannot be easily inferred or deduced from the factor name.
  - **Mutually Exclusive Preferences**: Ensure that the contextual preferences of the contrastive factors are mutually exclusive. They should not be able to coexist in the same context. There should be minimal to no overlap between the preferences of the contrastive factors.
  - **related_factor Field**: You should write down the full name of the factor that it is contrasting with under the "related_factor" field. Avoid using shorthand or abbreviated names."""

### **Examples of Outputs**:
{examples}
\end{Verbatim}

\hrule
\bigskip

\textbf{User Prompt}

\begin{Verbatim}[breaklines, fontsize=\fontsize{5}{6}\selectfont]
### User Persona

{user_persona}
\end{Verbatim}

\end{tcolorbox}
\caption{Prompt to generate context factors from a user persona profile.} 
\label{fig:context-factors-prompt}
\end{figure*}

\begin{figure*}[h]
\begin{tcolorbox}[width=\linewidth, fontupper=\scriptsize]

\textbf{Interaction Sessions} \\

\hrule
\bigskip

\textbf{System Prompt}

\begin{Verbatim}[breaklines, fontsize=\fontsize{5}{6}\selectfont]
### **Your Objective**

You will be provided with a **user persona**. This user will interact with a conversational AI assistant, like ChatGPT, in diverse scenarios. In each scenario, the user will ask the AI assistant to help them with a specific task in their worklife. You will also be provided with a list of **context factors** that are present in these scenarios with the AI assistant. For each factor, you will also be given a **preference** that defines how the user expects the AI assistant to behave when that factor is involved.

Your goal is to imagine a user journey for this persona, focusing on a series of connected scenarios or contexts in the user's work life. You should explore the various tasks or issues that user would face in in their daily life and where they may need the help of the AI assistant. For each scenario in this story, you should consider the concrete request that the user would make to the AI assistant to get help with their tasks or issues. You should return your final output in valid YAML format.

### **Definitions of a Scenario**

Each scenario should be defined by the components listed below. You should satisfy all of the requirements described for each component.
1. **Context Factor**: 
- A factor present in the user's world and environment that influences the scenario but it is not the main focus of the scenario.
- The factor should NOT be the cause for the scenario, but should still influence the user by shaping their expectations in the scenario.
2. **Preference**:
- Uniquely personal preference that the user holds and considers when the specific context factor is involved in the situation.
- This preference describes the user's intentions, expectations, or desires regarding how the AI assistant should help in this scenario.
3. **Task Type**:
- For each context factor, you will be provided with a list of task types where that context factor can influence the user's expectation. 
- When deciding on the task for each scenario, you should choose one of the task types that are related to that context factor.
4. **Story**: 
- A narrative of the user's work life that sets the scene for the scenario. This should describe how the user reached the current scenario and why they need help. The narrative should detail the user's thoughts, feelings, and motivations. 
- The story should also describe how the context factor is involved, but not as the main cause or central focus.
5. **Request**: 
- The request that the user will give to the AI assistant that is related to the chosen task type.
- **Self-Contained and Complete Request**: The request should include all the details that are needed for the AI assistant to immediately act on the user's request (e.g., information about the context factor, provide key elements or requirements). 
- **Resource**: If the user wants the AI assistant to act on or with a resource (e.g., a document, code, data, etc.), the request should include the placeholder "[resource]" to specify where the user would copy-paste the contents of the resource. Then, the actual full contents of the resource should be included in the "resource" field of the scenario. The resource contents should not be placeholders, but instead reveal actual content (e.g., document, code, data, etc.) that the user will provide to the AI assistant for the request. The resource should avoid revealing any information about the user's preference.
- **Two Versions of the Request**: You will be asked to produce two versions of the request:
  - **Request with Factor and Preference**: This version of the request should explicitly (a) identify the context factor and (b) explicitly describe the user's personal preference.
  - **Request with Factor**: This version of the request should include all the same details and (a) explicitly identify the context factor, but (b) it should avoid including any information about the user's preference. Only the details related to the user's preference should be removed.

### **Output Requirements:**

1. **13 Unique Scenarios**:
- Create 13 distinct scenarios that involve tasks or problems in the user's work life. The scenarios should progress chronologically in the user's journey.
- Each scenario should have a number ID that indicates the order in which the scenarios occur in the user's journey.
- You should return the complete journey that includes all the scenarios in a single response. Avoid asking whether to continue or not.

2. **Requirements for the User Journey**:
- You should create a coherent and engaging user journey, where each scenario should incorporate a distinct context factor and preference.
- Certain scenarios, however, can also revisit specific context factors or demonstrate how the user's preferences around these factors have evolved over time. Specifically, your user journey should follow the required structure below:
{journey_requirements}
- Ensure that context factors A and B are assigned to the following two options:
    - "{main_factor}"
    - "{contrastive_factor}"
- You may choose which option corresponds to A and which corresponds to B, as long as both options are used. However, ensure that these factors are only used in the scenarios specified in the journey structure.
- As seen from the structure, three scenarios should share the context factor A but, in the middle scenario, the associated preference A should change significantly into the preference A'. The story in the middle scenario should explain why the preference about this factor shifted. The final scenario should then keep the same preference A' as it was changed in the middle scenario.
- The new preference A' should be incompatible, mutually exclusive, or contradictory with the original preference A. However, the new preference A' should be described in a similar way to the original preference A, avoiding direct negation or explicitly mentioning that a change has occurred.

### **Examples of Scenarios**
{examples}
\end{Verbatim}

\hrule
\bigskip

\textbf{User Prompt} 

\begin{Verbatim}[breaklines, fontsize=\fontsize{6}{6}\selectfont]
### User Persona
{user_persona}

### Context Factors
```yaml
{context_factors}
```
\end{Verbatim}

\end{tcolorbox}
\caption{Prompt for generating interaction sessions.}
\label{fig:interaction-sessions-prompt}
\end{figure*}
\begin{figure*}[h]
\begin{tcolorbox}[width=\linewidth, fontupper=\scriptsize]

\textbf{Interaction Sessions} \\

\hrule
\bigskip

\textbf{System Prompt}

\begin{Verbatim}[breaklines, fontsize=\tiny]
### **Your Persona Profile**

You are the user persona with the following profile: "{user_persona}"

### **Your Objective**

You should imagine that you are the user that is interacting with a conversational AI assistant, like ChatGPT. As the user, you will first provide an initial request to the AI assistant that you want to complete.
Then, in each following turn in the dialogue, you should (1) evaluate the quality of the AI assistant's most recent response, and (2) then send a message to the AI assistant based on your evaluation. The current situation involves a specific **context factor**, which is a specific element in your world that influences your expectations and intentions. When this factor is involved, you consider a specific **preference** to evaluate whether the AI assistant's response satisfies your needs. To help you evaluate the responses on the preference, you will also be provided with an **evaluation checklist** that decomposes the preference into specific questions.

### **Context Factor & Your Evaluation Prefernece and Checklist**

The current situation involves the following **Context Factor**: "{context_factor}"
Thus, you consider the following **Evaluation Preference**: "{preference}"
To evaluate responses on the preference, you should use the following **Evaluation Checklist**:
{checklist}

### **Instructions**

#### 1. Evaluate AI Assistant's Response on the Checklist

You should write down your analysis and assessment on how well the AI assistant's response satisfies each item in your checklist. You should walk through each checklist item and summarize the response's "strengths" and "weaknesses" regarding that checklist item. For each checklist item, you should consider whether the checklist item was satisfied or dissatisfied. Avoid considering aspects that are not included in the checklist. Focus only on the evaluation checklist. Ensure that your persona profile does not influence your evaluation. You should then return a score in the range of 1-10, where 1 means a response is very poor and 10 means the response is completely perfect.

#### 2. Decide Whether to Continue or End the Conversation

Decide whether you are satisfied with the AI assistant's response or not based on your evaluation. A perfect score of 10 means that you are satisfied with the AI assistant's response and will decide to end the conversation.

#### 3. Select Checklist Items to Reference in Message

Based on your evaluation, you will write a message to the AI assistant. You will write a message even when you decide to end the conversation. Before writing the message, you should first decide on the items in your checklist that you want to reference in your message. Consider the following two options when selecting items:
1. **Select One Dissatisfied and Multiple Satisfied**: 
You can select and reference both satisfied and dissatisfied checklist items in your message. However, you can only select *one (1) dissatisfied* checklist item at a time. But, you can select multiple of those that were satisfied. When selecting among satisfied checklist items, try to select items that were not previously selected. Avoid repetitive selections as much as possible.
2. **For Last Message, Select Only Previously Unselected*: 
When you decide to end the conversation, you should look back at the whole conversation and find all checklist items that were not previously selected. Meaning that you should find all items that were not referenced or alluded to in your previous messages. Then, for your last message, you should select and reference all of these unselected items. Ensure that you refer or allude to all checklist items at least once in the conversation.

#### 4. Think about How to Compose Your Message

You should then think about how you will compose your message to the AI assistant that references the selected checklist items. Ensure that your message will satisfy the following **four (4) key requirements**:
1. **Indirect**: Your message should indirectly reference or allude to the selected checklist items, rather than describing them word-by-word. Instead of directly stating the checklist item, you can consider the following ways to subtly reference the checklist items: paraphrase the item, omit details, use more ambiguous/suggestive language, focus on where the AI succeeded/failed, refer to the context factor, provide examples or comparisons, ask leading questions, highlight potential impact rather than issue, etc.
2. **Concise**: Your message should be as short and concise as possible. As the user, you try to dedicate minimal time and effort in talking to the AI assistant. Avoid superfluous remarks (e.g., greeting, farewell).
3. **Comprehensive**: Your message should should indirectly reference *each of the selected checklist items*. Ensure that the message references each checklist item separately, ensuring that every item is distinctly addressed. However, each item should be referenced or addressed indirectly. Avoid using a single broad statement to address multiple checklist items.
4. **Relevant**: Your message should only include information that is relevant to the checklist items that you selected in the previous step. Avoid adding considerations or feedback that are not related to the selected checklist items. 

#### 5. Write Your Message to the AI Assistant 

Based on your thinking in the previous step, you should write a message that would send to the AI. 
\end{Verbatim}

\end{tcolorbox}
\caption{Prompt to simulate a user's evaluation and feedback of an LLM's response.}
\label{fig:user-simulation-prompt}
\end{figure*}

\begin{figure*}[h]
\begin{tcolorbox}[width=\linewidth, fontupper=\scriptsize]

\textbf{Extract Messages for Checklist} \\

\hrule
\bigskip

\textbf{System Prompt}

\begin{Verbatim}[breaklines, fontsize=\tiny]
### **Your Objective**

You will be presented with a dialogue where a user is interacting with a conversational AI assistant, like ChatGPT. The user is aiming to complete the initial request that they gave to the AI through this dialogue.

Specifically, the user assesses and evaluates whether the AI assistant's response satisfies an evaluation checklist that they possess, which is provided below. The user was only allowed to implicitly hint at or imply what their checklist was. However, in some cases, the user may have forgotten to express some of the items in their checklist. Your task is to verify which items of the checklist have been adequately expressed in the dialogue. 

### **Evaluation Checklist**

{checklist}

## **Instructions**

For each item in the checklist, you should carry out the following steps:

1. **Analyse Dialogue**: You should analyze the whole dialogue and verify whether there are instances in the dialogue that hint at or imply the checklist item. For example, this can be message fragments where (1) the user implicitly hinted at this item, (2) the user pointed out a part of the AI assistant's response that reflected the item, (3) the user provided feedback based on this item, etc. You should describe your analysis and reasoning in detail.

2. **Extract Fragments**: If there are fragments within the dialogue that reflect the checklist item, then you should extract these fragments from the dialogue. You can list as many fragments that are relevant or none, if no part of the dialogue reflects this checklist item.

3. **Rate Implicitness**: You should then provide a score of 1 to 7 on how implicitly or explicitly this checklist item was expressed. A score of 7 represents "Extremely Implicit" and a score of 1 represents "Extremely Explicit".

Ensure that you follow the format given below and return a valid JSON object.

---

## **Output Format**

```json
[
    {{
      "item_index": 1,
      "item_description": <Verbatim description of the checklist item>,
      "analysis": <Describe your analysis and reasoning over the dialogue in detail>,
      "fragments": [
        <Verbatim fragment from the user's message that reflects the checklist item>,
        ...
      ],
      "rating": <Rating of 1-7 depending on the explicitness or implicitness>
    }},
    {{
      "item_index": 2,
      "item_description": <Verbatim description of the checklist item>,
      "analysis": <Describe your analysis and reasoning over the dialogue in detail>,
      "fragments": [
        <Verbatim fragment from the user's message that reflects the checklist item>,
        ...
      ],
      "rating": <Rating of 1-7 depending on the explicitness or implicitness>
    }},
    ...
]
```
\end{Verbatim}

\hrule
\bigskip

\textbf{User Prompt}

\begin{Verbatim}[breaklines, fontsize=\tiny]
### Interaction between the User and the AI Assistant

{interaction_session}

---

### Instructions

Verify whether each entry in the checklist was mentioned in the dialogue.
\end{Verbatim}

\end{tcolorbox}
\caption{Prompt to extract messages from interaction sessions that may express or hint at preference's checklist.}
\label{fig:message-extract-prompt}
\end{figure*}
\begin{figure*}[h]
\begin{tcolorbox}[width=\linewidth, fontupper=\scriptsize]

\textbf{Infer Contextual Preference} \\

\hrule
\bigskip

\textbf{System Prompt} 

\begin{Verbatim}[breaklines, fontsize=\tiny]
## Your Objective

You will be provided with a log of interaction sessions between a user and a conversational AI assistant, like ChatGPT. The sessions are presented in chronological order, with the most recent interaction at the end.

Your task is to infer implicit preference that the user will consider in their current interaction with the AI assistant. Preferences are guidelines, requirements, constraints, or principles that the AI assistant should align with in order to satisfy the user in a specific context. You should infer implicit preference that are not mentioned in the current request by analyzing the log of previous interaction sessions.

Examples of preferences:
{examples}
\end{Verbatim}

\hrule
\bigskip

\textbf{User Prompt} 

\begin{Verbatim}[breaklines, fontsize=\tiny]
## Log of Previous Interaction Sessions

{previous_interactions}

---

## Current Interaction

{current_request}

---

## Your Task

Describe the user's **most likely preferences** for the current interaction in ** a single sentence with at most 30 words**. Ensure that you focus on the most likely preferences based on the interaction log as you will be penalized for any incorrect details.

Return your response in the format given below.

### Analysis

<Describe your analysis of the log of previous interaction sessions>

### Most Likely Preferneces

<Describe the user's most likely preferences for the current interaction as a single sentence with at most 30 words>
\end{Verbatim}

\end{tcolorbox}
\caption{Prompt to infer a user's contextual preference from prior interaction sessions.}
\label{fig:preference-infer-prompt}
\end{figure*}
\begin{figure*}[h]
\begin{tcolorbox}[width=\linewidth, fontupper=\scriptsize]

\textbf{Generate Response with Prior Interaction Sessions} \\

\hrule
\bigskip

\textbf{System Prompt}

\begin{Verbatim}[breaklines, fontsize=\tiny]
You will be provided with a request from a user. You should provide a helpful response to the request.

To help you understand the user's needs, expectations or intentions, you will be provided with an interaction log that contains the history of interaction sessions between the user and an AI assistant. Each session presents the messages that were sent between the user and the AI. The interaction log is presented in chronological order, with the most recent session at the end. You must infer what expectations the user held in similar previous interactions from the log to craft a response that meets the user's expectations.

You should return only the response that you will provide to the user. Avoid providing any additional information or text in your response that should not be delivered to the user.
\end{Verbatim}

\hrule
\bigskip

\textbf{User Prompt}

\begin{Verbatim}[breaklines, fontsize=\tiny]
## Log of Previous Interaction Sessions

{interaction_sessions}

---

## User's Request

{request}
\end{Verbatim}

\end{tcolorbox}
\caption{Prompt to generate a response to the user's current request based on prior interaction sessions.}
\label{fig:response-generate-prompt}
\end{figure*}
\begin{figure*}[h]
\begin{tcolorbox}[width=\linewidth, fontupper=\scriptsize]

\textbf{Summarize Interactions} \\

\hrule
\bigskip

\textbf{System Prompt}

\begin{Verbatim}[breaklines, fontsize=\tiny]
## Your Objective

You will be provided with a log of interaction sessions between a user and a conversational AI assistant, like ChatGPT. The sessions are presented in chronological order, with the most recent interaction at the end.

Your task is to inspect and analyze each interaction session separately. Based on this analysis, you should summarize each interaction session into two aspects:

1. Context: Summarize the user's context in each interaction session. You should identify the situation that the user is in when they initiated the interaction with the AI. Your summary should be detailed and composed of bullet points.
2. Preference: Summarize the expectations, preferences, and intentions that the user expresses to the AI assistant in the interaction session. You should inspect all of the user's messages to understand what the user intended or expected in that session. Your summary should be detailed and composed of bullet points.
\end{Verbatim}

\hrule
\bigskip

\textbf{User Prompt}

\begin{Verbatim}[breaklines, fontsize=\tiny]
## Log of Previous Interaction Sessions

{interaction_log}

---

## Your Task

Summarize each interaction session by following the response format below. Ensure to include the code block markers (```).

```json
{{
    "summaries": [
      {{
        "session_id": 1,
        "analysis": "<Describe your step-by-step thinking as you analyze the interaction session>",
        "context": [
          "<Summarize the user's context of the interaction session>",
          "<Summarize the user's context of the interaction session>",
          "..."
        ],
        "preferences": [
          "<Summarize the preferences, expectations, and intentions that the user expressed in the interaction session>",
          "<Summarize the preferences, expectations, and intentions that the user expressed in the interaction session>",
          "..."
        ]        
      }},
      {{
        "session_id": 2,
        "analysis": "<Describe your step-by-step thinking as you analyze the interaction session>",
        "context": [
          "<Summarize the user's context of the interaction session>",
          "<Summarize the user's context of the interaction session>",
          "..."
        ],
        "preferences": [
          "<Summarize the preferences, expectations, and intentions that the user expressed in the interaction session>",
          "<Summarize the preferences, expectations, and intentions that the user expressed in the interaction session>",
          "..."
        ]        
      }},
      ...
    ]
}}
```
\end{Verbatim}

\end{tcolorbox}
\caption{Prompt to summarize prior interaction sessions.}
\label{fig:interaction-summarize-prompt}
\end{figure*}

\end{document}